\documentclass{ar-1col-S2O}
\usepackage[numbers,compress]{natbib}
\usepackage[final]{changes}
\definechangesauthor[name={Divya Shanmugam}, color=blue]{SS}

\usepackage{url}
\begin{document}

\markboth{Shanmugam et al.}{Generative AI in Medicine}

\title{Generative AI in Medicine}

\author{{Divya Shanmugam$^1$, Monica Agrawal$^{2,3}$, Rajiv Movva$^1$, Irene Y. Chen$^{4,5}$, Marzyeh Ghassemi$^6$, Maia Jacobs$^{7,8}$, and Emma Pierson$^{1,9}$}
\affil{$^1$Department of Computer Science, Cornell Tech, New York, United States, 10044}
\affil{$^2$Department of Biostatistics and Bioinformatics, Duke University, Durham, United States, 27705}
\affil{$^3$Department of Computer Science, Duke University, Durham, United States, 27708}
\affil{$^4$Department of Computational Precision Health, UC Berkeley and UCSF, Berkeley, United States, 94709}
\affil{$^5$Department of Electrical Engineering and Computer Science / Berkeley AI Research, Berkeley, United States, 94709}
\affil{$^6$Department of Electrical Engineering and Computer Science/Institute for Medical Engineering and Science, Massachusetts Institute of Technology, Cambridge, United States, 02139}
\affil{$^7$Department of Computer Science, Northwestern University, Evanston, United States, 60208}
\affil{$^8$Department of Preventive Medicine, Northwestern University, Evanston, United States, 60208}
\affil{$^9$Department of Population Health Sciences, Weill Cornell Medical College, New York, United States, 10044. Email: emma.pierson@cornell.edu}}

\begin{abstract}
The increased capabilities of generative AI have dramatically expanded its possible use cases in medicine. We provide a comprehensive overview of generative AI use cases for clinicians, patients, clinical trial organizers, researchers, and trainees. We then discuss the many challenges -- including maintaining privacy and security, improving transparency and interpretability, upholding equity, and rigorously evaluating models -- which must be overcome to realize this potential, and the open research directions they give rise to. \end{abstract}

\maketitle

\section{Introduction}

Excitement about the promise of generative AI in medicine has inspired an explosion of new applications. Generative models have the potential to change how care is delivered \cite{tai2024ai, zakka2024almanac, garcia2024artificial,kamb2024, mannhardt2024impact}, the roles and responsibilities of care providers \cite{tierney2024ambient,marshall2020trialstreamer}, and the communication pathways between patients and providers \cite{feldman_scaling_2024,liu2024using}. 
Further upstream, generative models have shown promise in improving scientific discovery in medicine (through both clinical trials \citep{wang_autotrial_2023,white_clinidigest_2023} and observational research \citep{guo_automated_2024,ktena_generative_2024}) and facilitating medical education \citep{feldman_scaling_2024,bakkum_using_2024}. 
These developments are a direct result of technical advances in generative AI,
which have drastically increased the ability to generate realistic language and images,
and raise important questions about how to integrate generative models into medicine. 

Generative AI is the latest in a series of technical advances that have driven major shifts in medicine. Past significant advances include the adoption of electronic health records (EHRs); the integration of robotics into telesurgeries~\cite{dupont2021decade}; and the incorporation of predictive models and continuous monitoring as foundational infrastructure for new diagnostic tools~\cite{kononenko2001machine, razzak2020big}. But the introduction of new technologies into health settings inevitably introduces new challenges to overcome. For instance, the introduction of EHRs led to increases in data privacy concerns and data security breaches~\cite{kruse2017security, fernandez2013security}. And while the introduction of EHRs has led to significant reductions in medical errors and improvements in medical guideline adherence overall~\cite{campanella2016impact}, they have also introduced other types of errors~\cite{graber2019electronic}.
Similarly, continuous monitoring devices in healthcare settings have resulted in pervasive alert fatigue \cite{ancker2017effects}. Overall, the integration of technologies into medicine requires an iterative design process which addresses pitfalls and amplifies benefits \cite{rosenberg1982learning}. 

So, too, with generative AI. As generative models become a leading area of research and deployment in medicine, we provide a comprehensive review of the new applications they enable and the new challenges they create, with a particular focus on how users could interact with generative models. We first provide a brief overview of generative AI, detailing salient types of generative AI and how they fit into the broader landscape of machine learning in medicine. We next discuss the myriad use cases for generative AI in medicine, organized by potential users: clinicians ($\S$\ref{sec:use_cases_clinicians}), patients ($\S$\ref{sec:use_cases_patients}), trial organizers ($\S$\ref{sec:use_cases_trial_organizers}), researchers ($\S$\ref{sec:use_cases_researchers}), and trainees ($\S$\ref{sec:use_cases_trainees}). We then highlight the challenges ($\S$\ref{sec:risks_and_challenges}) that must be addressed to realize this potential and safely deploy generative models (including ensuring informed consent, protecting privacy, and improving transparency, among many other considerations) and discuss future directions for research throughout.

\subsection{Background on Generative AI}
\emph{Generative modeling} is a fundamental AI paradigm which stands in contrast to \emph{predictive modeling} (also called discriminative modeling): predictive models are given an input and seek to predict its label but do not attempt to model the input, whereas generative models do seek to model the input. For example, while a predictive model might be given a clinical note (the input) and try to predict whether the note indicates the presence of cancer (the label), a generative model would aim to model the distribution of clinical note text itself. The fact that generative models are trained to model the entire data distribution affords them the powerful ability to \emph{generate new data}: for example, to write new clinical notes. 

The basic generative modeling paradigm far predates the current surge of interest in generative AI. For example, classical generative modeling methods such as Markov chains have been used to model sequences of words for decades~\cite{bishop2006pattern}, and in theory could be used to write clinical notes. In practice, however, classical generative modeling methods do not generate sufficiently realistic content to be useful, especially on complex medical data. The current surge of interest is fueled by a drastic increase in generative modeling capabilities, which has been driven by scaling models to larger deep learning architectures and larger datasets \cite{kaplan_scaling_2020a}. 
These improvements have, as we describe, expanded the number of useful applications of generative AI models, and have spurred interest in applying these models to domains outside of core machine learning \cite{movva2024topics}.

We summarize three categories of generative models organized by the \emph{type of data} on which the model operates: (1) text, (2) images, or (3) text and images. For each category, we focus on the state-of-the-art models that are currently in use. While we limit our discussion to text and image data since these are most relevant to the use cases we subsequently discuss, generative models for other types of data (for example, physiological signals and molecular graphs) are an emerging area of clinical AI research \citep{abbaspourazad_large-scale_2023,beaini_towards_2023,mckeen_ecg-fm_2024}. For more comprehensive overviews, we refer the reader to \citep{bond-taylor_deep_2021}. 

For modeling text, \textbf{large language models (LLMs)} are the dominant approach, with substantial performance gains in recent years.
LLMs commonly use the transformer neural network architecture \citep{vaswani2017attention} to perform next word prediction: given a sequence of words, what is the most likely word that will come next? 
That is, for a context sequence $x_1, \cdots, x_n$, an LLM is trained to predict $p(x_{n+1} \mid x_1, \cdots, x_n)$ \cite{jurafsky2000speech}. 
What makes an LLM ``large'' is the size of its deep learning architecture, and the amount of data and compute used to train it; most language models in use today are considered LLMs.
Training an LLM usually consists of three phases: first, the LLM is \textit{pre-trained} on a large corpus of text scraped from the Internet; second, it is fine-tuned on \textit{instruction-following} examples, wherein user questions or instructions like ``Convert this discharge report into layperson's terms''~are followed by reasonable responses; third, the LLM is tailored with \textit{human feedback}, where humans choose which of two possible responses they prefer to capture fine-grained preferences \cite{ouyang_Training_2022}. Each of these phases can be tailored more specifically to medicine: some models are pre-trained on medical corpora, usually PubMed, in addition to or instead of the entire Internet \cite{bolton_biomedlm_2024, chen_meditron70b_2023}; some models are trained specifically to respond to medical questions using datasets like MedQA \cite{singhal_expertlevel_2023, jin_what_2021}; and there are emerging datasets of physician-written responses to medical queries to help align LLMs to medical best practice \cite{fleming_medalign_2023}.
While the use cases we will discuss primarily involve models trained on human language, models with similar transformer architectures can also be trained on other types of sequential biomedical data. For example, electronic health record models have been trained on sequences of ICD codes \cite{rasmy_medbert_2021, hill_chiron_2023}; protein models have been trained on sequences of amino acids~\cite{ferruz2022protgpt2}; and DNA models have been trained on nucleotide sequences~\cite{rives_Biological_2021, nguyen_sequence_2024}. 

For modeling images, \textbf{diffusion models}~\cite{yang_diffusion_2023} have recently become the method of choice, largely surpassing the previous generation's generative adversarial networks \cite{dhariwal_diffusion_2021}. 
Given an unlabeled training distribution of images, a diffusion model learns to \replaced{generate}{sample} new, synthetic images that look like images from the training distribution. 
To train a diffusion model, a real image $x_0$ is progressively corrupted to produce $x_t$, an image after $t$ corruption steps that looks like random noise. The model is trained to reconstruct the original clean $x_0$ from noise $x_t$ by learning the distribution $p(x_{i-1} \mid x_i)$. 
A well-trained diffusion model can then start from randomly-sampled noise, and produce a new image that is not in the training set but appears to have been drawn from the same distribution. 
Medical diffusion models have been trained on several different image types, such as chest X-rays, dermatoscopic images, and pathology slides \cite{kazerouni_diffusion_2023}.
To improve the biological validity of generated images, datasets and methods are being developed to fine-tune models using physician feedback \cite{sun_aligning_2023a}.
Synthetic image generation can be a useful augmentation technique in data-constrained settings, where supervised ML models may benefit from synthetic datapoints; recent evidence suggests this technique can help improve model robustness for pathology and radiology tasks~\cite{ktena_generative_2024}.

For tasks involving text \textit{and} images, there are two key types of generative models: \textbf{text-to-image (T2I)} and \textbf{vision-language models} (VLMs)\footnote{The latter are sometimes also referred to as image-to-text models.}.
T2I models accept a piece of \textit{text} as input, which is used by a text-conditioned diffusion model to generate a corresponding image as output.
These models consist of two components: a text encoder model (i.e., a transformer \cite{radford_learning_2021}), and a diffusion model which generates the image using the text encoding.
T2I models are pre-trained using general image caption datasets; they can then be fine-tuned for medicine, e.g.~using chest X-rays and corresponding radiology reports \cite{chambon_adapting_2022}. 
T2I models further expand the possibilities of synthetic images, for example by allowing researchers to generate training data for a specific patient pathology.
Relatedly, VLMs take in an \textit{image} as input and generate text involving the image as output \cite{li_BLIP2_2023}. 
VLMs consist of an image encoder model (e.g., a convolutional neural network or a vision transformer \cite{radford_learning_2021, li_BLIP2_2023}) and a large language model which can generate text based on the image encoding.
VLMs also require large image-text datasets, which can include image captions or reports, but also answers to visual questions, like ``Does this chest X-ray exhibit pleural effusion?'' \cite{thawakar_xraygpt_2024}.
They can be applied to tasks like question answering and report generation for pathology or radiology \cite{bazi_vision_2023, li_llavamed_2023}.

Each of these model classes have natural applications in medicine. Many clinical processes and decisions involve unstructured text (in the form of clinical notes, online health information, and treatment plans) and medical images. Moreover, images and text often appear in combination, as is the case with radiology reports. The next section elaborates on the potential to combine these generative modeling paradigms with existing data and processes in medicine.

\section{Use cases for generative AI in medicine}

\begin{figure*}
\centering
    \includegraphics[width=\textwidth]{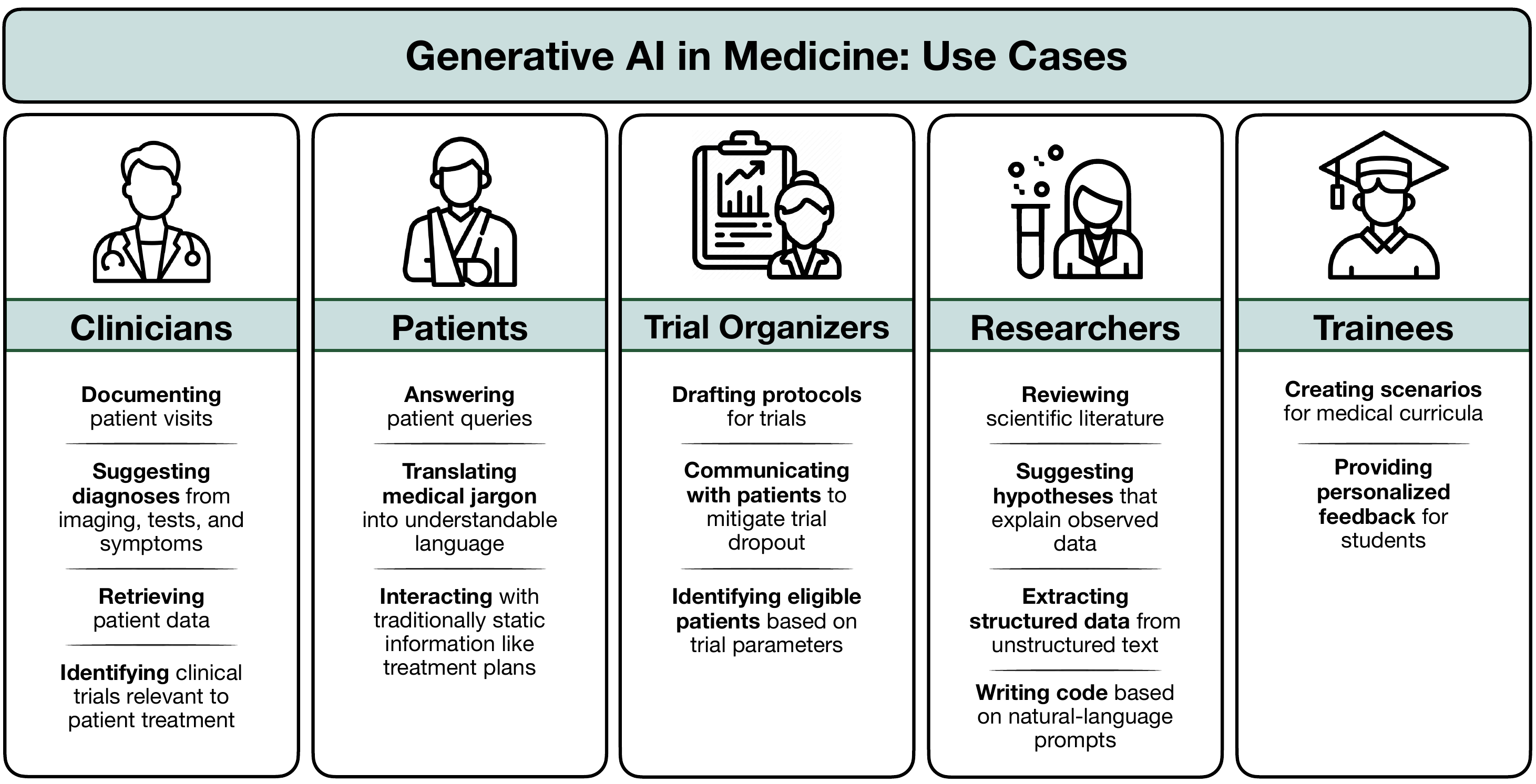}
    \caption{We highlight promising use cases of generative AI in medicine for five key constituent groups: clinicians, patients, trial organizers, researchers, and trainees.}
    \label{fig1}
\end{figure*}

The use cases for generative AI in medicine are numerous (Figure \ref{fig1}). We organize our discussion around key constituents in patient care: clinicians, patients, clinical trial organizers, researchers, and trainees. Within each role, we highlight core responsibilities that can be transformed by the introduction of generative AI. 

\subsection{Clinicians}
\label{sec:use_cases_clinicians}

Generative models have the potential to improve clinicians' ability to provide care in multiple ways: by assisting with writing and documentation (reducing administrative burden and physician burnout, which are major concerns~\citep{aiken_physician_2023}); assisting with diagnosis; retrieving patient data; and supporting evidence-based medicine. Each use case suggests that generative models can act as a useful tool for delivering care personalized to patient needs.

\subsubsection*{Assistance in writing} The amount of documentation in the electronic health record is a leading cause of physician burnout \cite{saag2019pajama}. Outside of their shifts, clinicians frequently spend significant time finishing clinical notes and communicating with patients; generative AI provides an opportunity to speed up both of these documentation processes. 

The earliest deployments of generative AI-driven writing assistance have been in ambient documentation, in which recordings of patient-provider communications are used to generate initial drafts of notes \cite{tierney2024ambient}. In initial pilots, providers and patients alike have indicated that the technology facilitates improved interactions, and providers have noted a decrease in time spent completing notes after their shifts.

In another example, the past few years have seen an uptick in provider-patient messaging through the electronic health record. While this has provided a straightforward path for patients to surface concerns and questions to their providers, providers have had to similarly spend increasing amounts of time responding \cite{saag2019pajama}. There have already been several pilots using large language models to respond to patient queries \cite{tai2024ai, garcia2024artificial}. Additional work has explored having large language models prompt \textit{patients} with follow-up questions automatically so that clinicians have full patient context with fewer back-and-forths \cite{liu2024using}. While clinicians report high usability of response drafting tools and decreased burnout, early results show that significant editing is still required, and effects on documentation time have yet to be seen.

Across both of these use cases, there are significant nuances which must be addressed for successful deployments. First, the clinical note writing process is not one-size-fits-all, as workflows differ significantly between specialties and clinicians. Second, significant effort is necessary to ensure that automation in the writing process does not lead to automation bias or decreased agency from clinicians \cite{heer2019agency}. Clinical notes often serve not only as documentation for future reference, but also as an active space for 
iteration and re-analysis
\cite{mamykina2012clinical} as, in many specialties, notes are not written at once, but in several sessions stretched over a patient's stay \cite{jiang2023conceptualizing}.

\subsubsection*{Assistance in diagnosis} 

Several works have assessed the efficacy of using generative models to provide diagnoses from patient information (e.g., medical imaging and lab test results). 
While current generative models do not identify the correct diagnosis with sufficient reliability to be used without clinician supervision~\cite{young_diagnostic_nodate,kanjee2023accuracy,levine_diagnostic_2023}, their generated differentials (i.e. sets of diagnoses) often contain the correct diagnosis and could be useful as a tool to \emph{augment} clinicians by expanding the set of diagnoses a clinician considers~\citep{rios-hoyo_evaluation_2024}. 
Such a tool could be particularly useful in contexts where identifying candidate diagnoses is difficult (for example, rare diseases \citep{olmo_assessing_2024} or challenging clinical cases~\cite{kanjee_accuracy_2023}).
\cite{levine_diagnostic_2023} find that a large language model includes the correct diagnosis in its differential in 88\% of cases, nearing the diagnostic performance of clinicians, who include the correct diagnosis in 96\% of cases. Other works have studied the accuracy of generative model diagnosis given both images and text \citep{zhou_pre-trained_2024,thawkar_xraygpt_2023,moor_med-flamingo_2023,lin_medical_2023}; \citep{zhou_pre-trained_2024}, for example, demonstrate how a vision-language model can identify the correct skin condition, given an image and a text prompt, in 80\% of cases. 
Overall, diagnostic performance depends on both the task and generative model \citep{reese_limitations_2024,zhou_pre-trained_2024,hager_evaluation_2024}, and while significant progress must be made for generative models to provide diagnostic assistance to clinicians, the initial results are promising. Ultimately, such a tool could be used in service of precision medicine by providing diagnostic assistance personalized to a patient's medical history.

\subsubsection*{Information retrieval of patient data}
Clinical practice requires a significant amount of \textit{synthesis} of a patient's past health narrative (e.g., understanding past medications, lab trends, and diagnoses) with their current clinical state \cite{mamykina2012clinical}. In current EHRs, clinicians often have trouble finding the relevant information needed for contextualized clinical care, both due to the fragmentation of EHRs and the amount of information trapped in free-text notes \cite{ahmed2011effect}. Finding relevant labs or medical history can often involve time-consuming and disjointed navigation through different sections of the electronic health record.  Retrieval-augmented generation (RAG), in which generative models retrieve information from an external database and use it to inform the language they generate, provides a possible avenue for clinicians to  conveniently surface data in a single unified process, powered by natural language \cite{zakka2024almanac}. RAG combines the flexibility of generative AI with more classical information retrieval systems; for example, a clinician could query a generative model with ``relevant family history for chest pain" or the ``result of the patient's last colonoscopy" to retrieve information across a patient's EHR. Going even further, the model could proactively surface information likely to be relevant, based on the current stage of a visit. Here, open questions remain on what information to automatically surface, when to do so, and how to best display it \cite{murray2021medknowts, jiang2023conceptualizing}. Previously, researchers have found it useful to draw on past interaction patterns in the electronic health record in order to inform interface design \cite{zheng2009interface}. 

\subsubsection*{Evidence-based medicine}
Evidence-based medicine requires bringing new findings from clinical research to the bedside, but it can be nearly impossible for clinicians to keep up with the pace of clinical research \cite{sackett1995need, bastian2010seventy}. Large language models can make it easier to organize and query clinical trial information at the point-of-care. Existing explorations have included NLP-generated databases of cleaner, searchable clinical trial information and natural language interfaces for interacting with clinical guidelines more directly
\cite{marshall2020trialstreamer, zakka_almanac_2023,lee_seetrials_2024}.

\subsection{Patients} 
\label{sec:use_cases_patients}
Patients often express a desire for increased involvement in their care, e.g. in the form of shared decision making or increased access to information \cite{chewning2012patient}. Below, we describe a few ways generative AI could impact the patient experience. We emphasize to the machine learning community the necessity to embrace \textit{participatory design}, where proposed tools or interfaces (for example, interfaces to surface health information) are built in collaboration with patients and center stakeholder perspectives, 
as has been done in informatics and human-computer interaction \cite{adler2022developing, noack2021designing, danieli2021conversational, martin2019exploring}.

\subsubsection*{Searching for Online Health Information}
Patients often leverage the Internet to help answer their health questions, particularly when they have limited access to care \cite{amante2015access}. For those with access, benefits have included increased ability to participate in decision making, more informative questions during episodes of care, and improved adherence to instructions \cite{thapa2021influence}. Recent surveys show that generative AI has already seen significant adoption for information seeking between appointments \citep{vanessa_choy_can_2024,alex_montero_kff_nodate}.  Unlike existing search engines, generative AI enables patients to pose more specific queries and ask follow up questions, allowing for more tailored responses and conversational searching where questions and responses can build upon one another.
However, the current generation of models can provide plausible but inaccurate information, making it challenging for patients to discern when to trust model outputs~\cite{hersh2024search}.  Patients are still faced with the challenge of discerning if the information is accurate, and current generative AI chatbots provide few references. We expand on these issues in later sections.

\subsubsection*{Increasing Patient Engagement}
Increased patient engagement can enable patients to better understand their own health conditions and care plan. This engagement can also lead to an increase in patient-reported outcomes (e.g., self-tracking and sharing of symptom burden); this in turn can enable clinicians to better understand their patients' conditions in between visits \cite{grossman2018leveraging}. One approach to increased patient engagement is the use of patient portals; however, utilization of patient portals remains low \cite{zhao2017barriers}. Given their potential, there have been several long-running suggestions for how to increase utility of patient portals that are now more feasible with generative AI \cite{grossman2018implementation, grossman2019interventions, warren2019working}. As an example, clinical notes contain valuable information for patients but were not written with patients as the intended audience: they are filled with jargon, and patients cite the difficulty of medical jargon as a major barrier to comprehension \cite{kambhamettu2024explainable}. Recent work has explored using generative models to \textit{translate} clinical notes into patient-friendly language and visualizations, with the opportunity to personalize to information needs of patients and how they want that information presented \cite{kamb2024, mannhardt2024impact,luo_rexplain_2024}. The opportunity of patient-friendly text simplification extends past clinical notes alone to other facets of health literacy, e.g., medical literature and patient consent forms \cite{basumedeasi, mirza2024using}.

\subsection{Trial Organizers}
\label{sec:use_cases_trial_organizers}

Clinical trials provide critical evidence to update and improve clinical practice. 
Conducting these trials, however, remains challenging: 
only 20\% of clinical trials in the United States complete within the planned timeframe \citep{shadbolt_analysis_2023}, and of those that do, only a small fraction are published \citep{ross_time_2013,zarin_update_2017}. 
The difficulty of clinical trial design is in part due to the complexity of interaction and documentation involved; trials can fail due to incorrectly designed protocols, insufficient participant registration, or high patient dropout \citep{getz_impact_2016,fogel_factors_2018}. Generative interfaces present the potential to rework key components of this pipeline.

\subsubsection*{Protocol Design} Clinical trial design begins with the creation of a protocol, which collates existing research, study aims, and regulatory requirements into concrete steps detailing how the trial will proceed. Protocol creation requires significant manual effort \citep{ghim_transforming_2023}, and existing work has illustrated the value in using generative models (specifically, large language models) to expedite the process \citep{wang_autotrial_2023,white_clinidigest_2023,ghim_transforming_2023,park_criteria2query_2024,lai_assessing_2024,datta_autocriteria_2024}. Several works center on the generation and evaluation of exclusion and inclusion criteria \citep{yuan_criteria2query_2019,park_criteria2query_2024,wang_autotrial_2023,datta_autocriteria_2024}, while others propose the use of large language models to retrieve relevant past clinical trials to inform the construction of a new protocol 
\citep{white_clinidigest_2023,ghim_transforming_2023}. \cite{lai_assessing_2024} employs large language models to evaluate protocols for bias automatically, while \cite{wang_autotrial_2023} uses large language models to generate trial inclusion and exclusion criteria based on details of the setup expressed in natural language. Together, the literature thus far highlights the potential for generative interfaces to reduce the time required to construct a successful protocol.

\subsubsection*{Participant Recruitment \& Retention} Equipped with a protocol, clinical trial designers must recruit a sufficient number of participants to conduct the trial. Generative interfaces can be used to identify suitable trial participants by parsing both the criteria and patient history \citep{hamer_improving_2023,wornow_zero-shot_2024,jin_matching_2024,beattie_utilizing_nodate}. Early results suggest that large language models can reduce the number of eligibility criteria a clinician must manually check to assess eligibility by 90\% \cite{hamer_improving_2023} and reduce the time it takes to assess eligibility by 42\% \cite{jin_matching_2024}.

As the trial progresses, patient dropout can threaten its validity and success. Patients drop out of clinical trials for a multitude of reasons, one of which is poor communication with trial recruiters and clinicians \citep{mccann_recruitment_2013,skea_exploring_2019}. Patient dropout is particularly salient to decentralized clinical trials, which are conducted at non-traditional sites (e.g. a patient's home, or a local clinic) and are known to recruit more diverse patient populations compared to their standard counterparts \citep{goodson_opportunities_2022}. \citep{thomas_artificial_2022} propose the use of AI to improve patient engagement in these settings, which could include chat interfaces to answer trial-related patient questions as they arise. Such an interface could effectively and efficiently address patient concerns and misconceptions about a trial, including the potential for adverse events \citep{zhou_cancer_2019} or the importance of trial participation regardless of treatment outcome \citep{skea_exploring_2019}.

\subsection{Researchers}
\label{sec:use_cases_researchers}

Designing a useful medical study is a time-consuming process. Researchers comb through large bodies of literature, across multiple disciplines, to identify open questions and understand the status quo in different application areas. Later stages of research, including dataset construction, hypothesis generation, and code generation are no easier. Here, we highlight how generative models can help alleviate significant manual effort at each step. 

\subsubsection*{Literature review}  Efforts to collate existing literature into a coherent research question precede any effort to execute the study. Large language models have been shown to be a promising tool for problem generation through automated systematic literature reviews  \cite{guo_automated_2024,dennstadt_title_2024}. Here, a generative interface can allow a researcher to automatically assemble a clinical review by querying thousands of clinical abstracts, substantially reducing the effort required to perform a systematic review. \cite{guo_automated_2024} demonstrate how a popular large language model (GPT-4) can identify relevant papers at 91\% accuracy compared to human evaluation, with the ability to justify the inclusion and exclusion of particular papers. 

\subsubsection*{Dataset construction} Generative AI can improve the quality, quantity, and diversity of datasets in medicine \citep{chen_synthetic_2021,ktena_generative_2024,khosravi_synthetically_2024}. To produce such datasets, generative models can be used in two ways: to \emph{generate completely synthetic data}, or to \emph{extract structured information} from existing unstructured data. 

Using generative models to create synthetic data has shown promise in compensating for gaps in existing datasets \citep{ktena_generative_2024,das_conditional_2021,ive_generation_2020}. \citep{das_conditional_2021} show that synthetically generated images can be used to improve a machine learning model's ability to detect COVID-19. Beyond improvements to accuracy, \citep{ktena_generative_2024} illustrated how augmenting training data using generative models can improve the fairness of the resulting diagnostic classifier. These findings hold in the context of natural language; \cite{ive_generation_2020} demonstrate how diagnostic classifiers trained on generated text perform comparably to those trained on real datasets of the same size, suggesting that synthetic data is a promising approach to addressing data limitations.

Generative AI can also be used to extract structured information from semi-structured data or to label existing data, across both natural language and imaging. Research on health equity, for example, relies on structured fields describing patient demographics to assess the severity of health disparities. Large language models can be used to extract demographic data from unstructured text (e.g. clinical notes) \citep{pierson2023use}, thus enabling comparisons of health outcomes between demographic groups.
Similarly, several generative AI tools have also been developed to measure morphological features from large cohorts of histopathology images using natural language prompts \cite{sun_pathasst_2024,lu_multimodal_2024,huang_visual-language_2023}. Generative models can also be used to alleviate the burden of data annotations by providing synthetic labels (i.e. predictions of ground truth) for  unlabeled examples \cite{wang_scientific_2023}; for example, one could use a generative model to suggest candidate segmentations of medical images \citep{wong_scribbleprompt_2024}.

\subsubsection*{Hypothesis generation} Given a dataset, generative models can be used to surface hypotheses \cite{zhou_hypothesis_2024,zhong_goal_2023,pham_topicgpt_2024,kamienny2022end}.  
\cite{zhou_hypothesis_2024} examine the use of generative models to produce natural language hypotheses (i.e. ``customers tend to buy shoes that match the color of their shirt"). They find that the resulting hypotheses confirm expected relationships, provide new insights, and outperforms supervised baselines.
\cite{zhong_goal_2023} frame the discovery of drug-specific side effects as a task for a generative model; given patient feedback for different drugs, the generative model is tasked with describing differences between drug-specific patient feedback in natural language.

\subsubsection*{Code generation} Generative interfaces can also be used to write code based on natural language prompts, which could lower the barrier for researchers to perform quantitative analysis of large-scale datasets \cite{tayebi_arasteh_large_2024}. \citep{tayebi_arasteh_large_2024} demonstrate that GPT-4 is capable of autonomously producing code to train models for disease screening and diagnosis. Collaboration with a code assistant has been shown to improve programming productivity \cite{mozannar_realhumaneval_2024}, and could help facilitate quantitative analysis of increasingly large observational health datasets. 

\subsection{Trainees}
\label{sec:use_cases_trainees}

Generative interfaces are already in widespread use by medical trainees \cite{biri_assessing_nodate}. The rapid uptake of these tools among trainees suggests the potential for generative interfaces to significantly transform medical education. Two promising applications are the use of generative interfaces to create practice clinical scenarios and provide feedback that targets student-specific areas for improvement.

\subsubsection*{Case creation} 
 Designing realistic clinical scenarios to test understanding is a critical yet difficult task. Those in charge of clinical curriculum design could use large language models to generate compelling multiple choice questions, as \cite{grigorian_implications_2023} have demonstrated in the context of surgical education. Doing so could also address the lack of diversity in clinical vignettes \cite{lee_race_2022}, and allow the generation of problems that better reflect patient populations trainees are likely to interact with \citep{bakkum_using_2024,benoit2023chatgpt,tejani_artificial_2023}. \cite{bakkum_using_2024}, for example, develop a tool to use large language models to produce 30 distinct cases in under an hour (including manual human review). Simulated patients could also be used to simulate real-world interaction. For example, the process of collecting patient history could be taught through a generative interface, in which a large language model responds to a medical student's requests for information based on a synthetic patient profile. \cite{holderried_generative_2024} have shown that such a set-up is well-received by medical students, and that more than 97\% of the generated answers were deemed clinically plausible. Each of these uses cases presents an opportunity to reduce the resources required to train medical students to deliver care. 

\subsubsection*{Providing personalized feedback} 

Generative interfaces can also provide tailored feedback to students \cite{dai_can_2023}. \citep{dai_can_2023} show that large language models can provide students with more coherent, process-oriented feedback compared to human instructors, while retaining high agreement with human-generated feedback. There is significant opportunity to apply these findings in medicine; for example, one could use large language models to provide feedback on a medical student's efforts to communicate health information to patients. 
Indeed, \citep{feldman_scaling_2024} showed that when generative models are used to provide feedback on clinical notes, the resulting notes are more complete, concise, and correct across four distinct specialties. 

Feedback for trainees can take many forms; in surgery, for example, non-generative AI is already being used to provide automated assessments of surgical procedures in simulated environments \citep{belmar_artificial_2023}, which helps trainees safely gain familiarity with procedures. 
\citep{zhao_surgical_2022} develop a generative model to output the optimal surgical path and show that real-time guidance leads to significant improvements along multiple surgery-specific performance metrics, including the number of attempts required to complete the surgery and risk of tissue damage.

\section{Challenges and directions for future work}
\label{sec:risks_and_challenges}

\begin{figure*}
\centering
    \includegraphics[width=\textwidth]{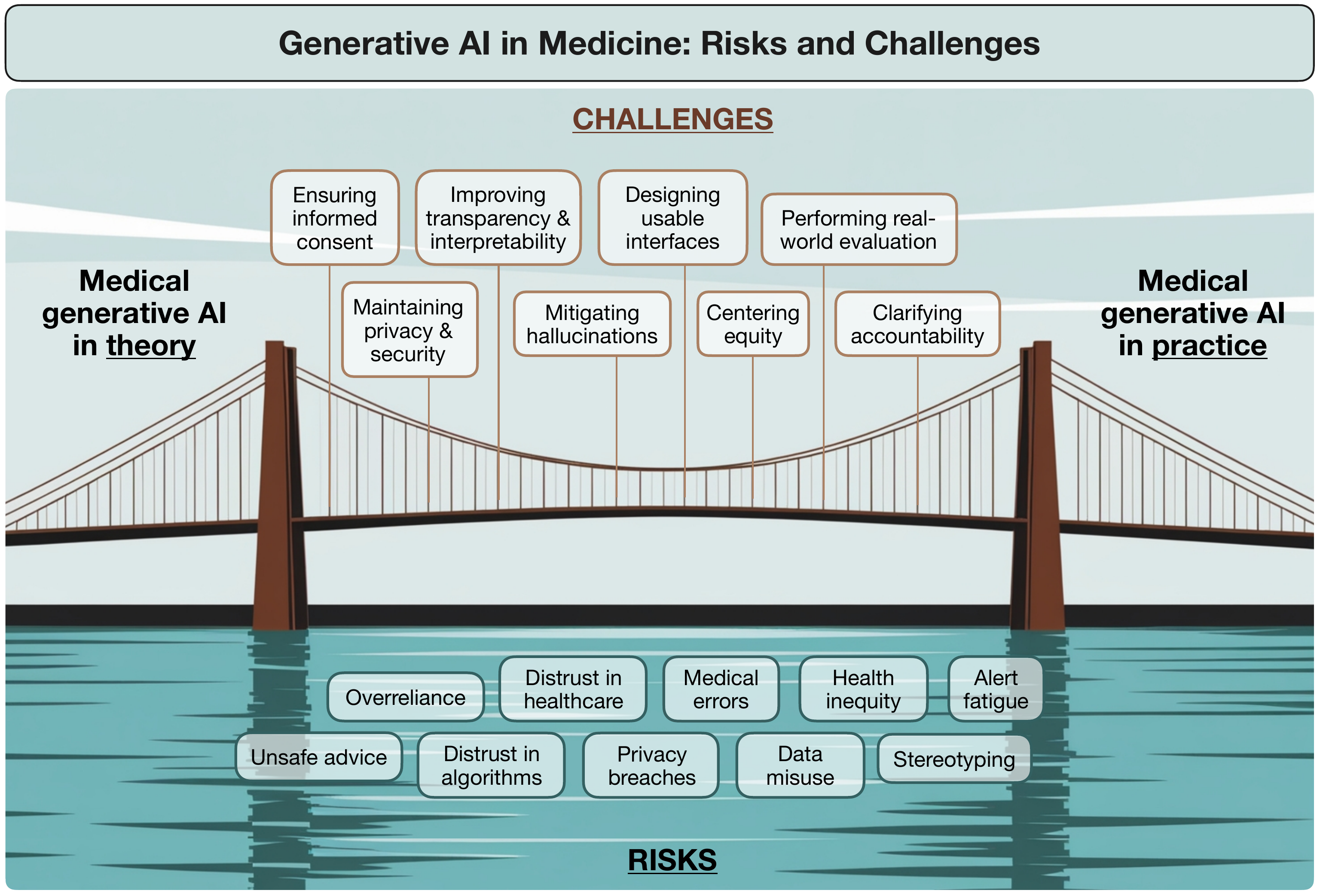}
    \caption{Bridging the gap between generative models in theory and practice will require addressing key challenges to mitigate risks and maximize benefits.}
    \label{fig2}
\end{figure*}

Fully realizing the opportunities to apply generative AI in healthcare requires significant progress on a number of challenges we describe below, and illustrate in Figure \ref{fig2}. Generative AI interfaces are far from perfect, and we are only beginning to understand the impacts of such interfaces on clinical decision-making. We have already seen examples of the potential harms such interfaces have caused that warrant attention. For example, \cite{hager_evaluation_2024} demonstrate how large language models fail to follow diagnostic guidelines up to 36\% of the time in an evaluation across realistic patient cases. Below, we enumerate challenges we foresee as critical to address as the intersection between generative interfaces and healthcare evolves, and discuss future directions for research throughout. 


\subsection{Ensuring informed consent}

Informed consent is a foundational principle of medical ethics which states that a patient must have access to sufficient information about a medical procedure (including risks, benefits, and alternatives) ~\cite{kirby1983informed,riddick2003code,del2005informed,pierson2022patients}. Achieving informed consent when using AI models raises new challenges which are a topic of active discussion~\cite{astromske2021ethical}: for example, how do we provide patients with comprehensive, accurate, and understandable information about complex models whose behavior is not fully understood even by their own creators (let alone the clinicians using them)? How do we ensure patients are consenting to the use of their data if it is used in model training? These issues similarly apply  to generative models, which also raise new challenges~\cite{wilcox2023ai,garcia2023ethical}. For example, when asked for their concerns about the use of generative AI models to transcribe and summarize patient-clinician conversations, providers expressed concerns 
about whether patients could meaningfully consent to the collection of this data~\cite{wilcox2023ai}. Similarly, the use of generative models in chatbots that interact with patients raises new concerns about informed consent~\cite{garcia2023ethical}. When chatbots and other generative AI tools are implemented in care practices, patients need to be given the option to decline the use of the models in their care and the use of their data. Patients must be provided with clear information that they are interacting with a chatbot, who the chatbot's creators are, and the uses and limitations of the technology. 

Contrasting with these concerns, generative models also show promise for \emph{improving} the informed consent process~\cite{decker2023large} by making consent forms easier to understand. A study comparing LLM-generated consent forms to those created by five surgeons for six common medical procedures found that the LLM-created forms were more readable and accurate than those created by surgeons. In this way, LLMs can help advance equity in medicine, through the creation of consent forms that are more accessible to a broader audience. Current medical consent forms are often written at a high reading level and describe complex procedures \cite{burks2019health}. As a result, many people, especially non-native English speakers and individuals with low literacy, are at increased risk of consenting to research and medical procedures without being fully informed \cite{simon2003groups}. While LLMs offer promise in improving consent documentation readability, more work is needed to ensure consent documents developed by LLMs contain comprehensive information \cite{raimann2024evaluation}.

\subsection{Maintaining privacy and security}

The use of generative models in medicine raises substantial privacy and security challenges~\cite{chen2024generative}, given the sensitivity and legal protections of medical data. One challenge is that generative models perform best (and are more likely to generalize) when trained on large, multi-institutional datasets, raising the question of how to share data across multiple institutions in a privacy-preserving way. Technical approaches like federated learning~\cite{bai2021advancing,ali2022federated,xu2021federated} offer one approach to this, although recent work has indicated that privacy violations are still possible in this setting \cite{geiping_inverting_2020,so_securing_2023,huang_evaluating_2021}. The creation of large de-identified datasets which can be securely shared with researchers~\cite{mullainathan2022solving,johnson2023mimic} is also an important catalyst for generative model research which institutions and policymakers should facilitate. 

After training generative models, a second challenge is \emph{deploying} models trained on sensitive data in a secure and private way. Past work has demonstrated that generative models can leak sensitive data by memorizing their training datasets and revealing private data in response to adversarially crafted prompts~\cite{barrett2023identifying,el2022impossible,carlini2022quantifying,huang2022large,carlini2021extracting}. Past work also suggests that these problems may grow worse as models continue to scale because larger models possess greater capacity to memorize the training data~\cite{carlini2022quantifying}. Mitigating these challenges remains an active area of research which is essential for safely deploying models trained on sensitive health data. 

Finally, and more optimistically, generative AI also holds potential for sharing data in a more privacy-preserving way, via generation of synthetic datasets which mimic properties of a real dataset while preserving patient privacy~\citep{choi_generating_2018,ghosheh2024survey,loong_disclosure_2013} (see $\S$\ref{sec:use_cases_researchers} for further discussion of synthetic datasets). For example, \citep{choi_generating_2018} demonstrate how synthetic patient records produced by a generative model can be substituted for real data at no loss of performance, with significant improvements to patient privacy.

\subsection{Improving transparency and interpretability}

Modern AI models are opaque for a number of reasons, and generative models are no exception. A first major challenge is a lack of \emph{transparency}: basic details of generative models are often not disclosed, including training data, training methods, model architecture, capabilities, limitations, and risks~\cite{bommasani2023foundation}. Lack of transparency causes several harms~\cite{bommasani2023foundation}: it makes it more challenging for policymakers to regulate models; for users to assess when they will perform reliably; and for researchers to innovate on them. Consolidation around a small number of closed models risks heightening the lack of transparency~\cite{movva2024topics}. The sensitivity of health datasets, which often cannot be publicly released, also makes achieving transparency more challenging. A 2023 review of widely used generative models scored them on 100 granular transparency indicators and found they averaged only 37 out of 100~\cite{bommasani2023foundation}, though the average score had improved to 58 out of 100 when the review was conducted in May 2024~\cite{bommasani2024foundationtransparency}, suggesting that transparency can be improved and that systematic reviews of the ecosystem are helpful. 

A related, but distinct, challenge is \emph{interpretability}: even if all details of a model are fully disclosed, understanding \emph{why} the model gives the output it does can be extremely difficult. Without understanding why a model produces a particular output, it is difficult to know whether to trust the model, and when it will fail. For example, healthcare models have been known to rely on spurious features to make predictions, and without knowing what features a model is using, these failure modes are difficult to identify~\cite{winkler2019association,oakden2020hidden,zech2018confounding}. Interpretability challenges are not unique to generative models, but occur with many modern AI models, including other deep learning architectures ~\cite{gilpin2018explaining,stiglic2020interpretability}. In general, interpretability methods (also known as explainable AI methods) have seen mixed success~\cite{ghassemi2021false,bilodeau2024impossibility}; different methods can yield very different answers, and those answers may be misleading. 

Similar interpretability challenges occur in the context of modern generative models, which can have billions or trillions of parameters, encoding non-linear, highly complex functions of the input data which are extremely challenging to understand or describe in a human-interpretable way~\cite{zhao2024opening}. While language-based generative models can provide plausible-sounding explanations for their reasoning~\cite{singh2024rethinking}, seemingly improving interpretability, those explanations are not necessarily accurate~\cite{agarwal2024faithfulness}. In general, the capabilities of generative models are currently advancing considerably more quickly than our ability to explain how they achieve those capabilities, which is concerning especially in high-stakes domains like healthcare. 

Improving interpretability of generative models remains an active research area; proposed approaches include \emph{local explanations}, which explain a single output from an generative model and \emph{global explanations} which explain a model's behavior as a whole~\cite{singh2024rethinking}. In a qualitative analysis of local explanations for a vision-language model applied to pathology images, 
\cite{u_vision-language_2023} find that the interpretations align with clinically known disease characteristics. 
Although the fidelity of such explanations is context-dependent, they remain a key ingredient in improving the transparency of generative models. 
Another recent line of work seeks to train generative models that are interpretable by design, by training models on paired images and text so the model can provide natural language annotations of generated images; \cite{kim2024transparent} apply this approach to dermatology data, and find that the models can accurately annotate images, as verified by dermatologists. A final way to address interpretability challenges is simply \emph{rigorous evaluation} of a model across a range of settings: even if it is not possible to understand exactly how a model produces its outputs, one can verify that they are reliably accurate. 

\subsection{Mitigating hallucinations}

Recent work has shown that generative models sometimes output medical information that is incorrect \cite{chen_use_2023} or hallucinated \cite{ji_survey_2023, lee2023benefits}. Hallucinated outputs in high-stakes medical settings can be dangerous: for example, they can harm patients without clinician access who rely on LLM-generated outputs for medical advice. Other work shows that LLM outputs can be hard to understand or non-actionable \cite{pan_assessment_2023}, which, while not directly harmful, may undermine widespread usability, especially for underserved patients. 

One promising approach to reducing generative models' propensity to hallucinate is \emph{retrieval-augmented generation} (RAG). As discussed above (\S\ref{sec:use_cases_clinicians}), RAG integrates traditional approaches towards search and information retrieval into generative models: specifically, by retrieving information from a verified knowledge base to guide the text a language model generates. RAG has been shown to reduce the extent to which large language models hallucinate \citep{shuster_retrieval_2021} and can also make a model's information sources more transparent, increasing users' ability to assess their reliability. 
While RAG and other methodological developments continue to mitigate hallucinations~\cite{zakka_almanac_2023,agrawal2022large,gero_self-verification_2023}, it's essential to continue validating accuracy and readability in each use case, especially more difficult or error-prone ones (e.g.,~as done in \cite{singhal_towards_2023}). 
We further discuss challenges in evaluating LLMs below (\S \ref{sec:evaluation}). In addition to technical methods and audits, regulatory oversight of generative models will also help mitigate the harms of inaccuracies \cite{mesko_imperative_2023}. 
Policymakers can, for example, encourage greater transparency in model development by mandating disclosure of adverse events and of important details which are currently often not disclosed, including training datasets, architectures, limitations, and biases.

\subsection{Designing usable interfaces}
Pioneers in human-computer interaction have called generative AI “the first new interaction model in more than 60 years” \cite{jakob_nielsen_ai_2023}. When interacting with generative AI, users are now able to tell the computer their desired intent (e.g. create a summary of melanoma for patients that includes symptoms, treatment options, and management strategies), rather than the exact actions they want the computer to take \cite{sai2024generative}. Thus, generative AI interactions can be efficient and low-burden \cite{mulia2023usability}, and offer a new way to design and develop novel health interventions \cite{giunti2024cocreating}. However, this new interaction paradigm brings usability challenges that have yet to be addressed. Numerous studies have shown that creating accurate and useful prompts for a generative AI platform is difficult for end users \cite{tankelevitch2024metacognitive, dang2022prompt, sun2022investigating, subramonyam2024bridging, zamfirescu2023johnny}. Once a prompt is provided, end users then face the additional challenge of interpreting and evaluating the output \cite{subramonyam2024bridging, abbasian2024foundation}. As we have seen with previous AI technologies, when end-users are unable to accurately evaluate the output of a model, they can become overreliant or make erroneous decisions \cite{vasconcelos2023explanations, kostick2022ai, wysocki2023assessing, jacobs2021machine}. To address these usability challenges, \cite{weisz2024design} recommend adopting user experience principles to guide the design of these systems. More work is needed to establish best practices for both the design and end-user evaluations of generative AI systems.

AI tools often face many obstacles to widespread adoption~\cite{schaefer2016meta,mertz2015annoying,greenes2018clinical,wright2016analysis} leading to limited health impact~\cite{khan2019improving,mann2020impact}. Given similarities of generative interfaces with prior AI tools, we expect similar challenges to arise. First, healthcare professionals may be hesitant or skeptical about new generative AI interfaces, making them resistant to change. In studies which simulated clinical settings with patients, research has found that provider experience levels~\cite{gaube2021ai,gaube2023non}, interface style~\cite{jacobs2021designing}, and time pressure~\cite{jacobs2021designing} may all affect adoption likelihood. Second, training and education are crucial to ensuring that healthcare professionals can best leverage these new technologies, which can be costly and time-consuming. Exposure through formal education or prior experience with similar AI interfaces can make healthcare professionals more comfortable with AI tools, leading to a higher rate in adoption~\cite{henry2022factors}. Educating users on the strengths and weaknesses of commonly used generative interfaces 
has been shown to improve qualities of human-AI collaboration including accuracy \cite{cabrera_improving_2023} and reliability \cite{bansal_beyond_2019}, and we expect these findings to hold true in medicine.
Lastly, deployment of generative interfaces may require an initial investment in cost and resource allocation. While preliminary studies have shown that generative interfaces can save time — e.g., initial estimates show that it may save nurses around 30 seconds per generated message~\cite{epicshareSavesNurses}— the potentially large upfront cost remains a key concern to active adoption. 

\subsection{Centering equity}

Abundant previous work has demonstrated how biases in medical datasets can propagate into AI models~\cite{obermeyer2019dissecting,gichoya2022ai,daneshjou2022disparities,gervasi2022potential,zink2023race,chen2021ethical,pierson2020assessing,wiens2019no,pierson2024accuracy,seyyed2021underdiagnosis,movva2023coarse,zou2023implications,balachandar2023domain,ferryman2023considering,shanmugam2024quantifying,zink2020fair,vyas2020hidden,mullainathan2021inequity,pierson2021algorithmic,obermeyer2021algorithmic,diao2024implications,diao2024projected}. It is thus unsurprising that the use of generative models in medicine creates several equity-related challenges~\cite{pfohl2024toolbox}. Research indicates that larger models do not, on their own, necessarily resolve equity concerns; indeed, larger models have been shown to exhibit more covert forms of bias (i.e. prejudice against certain dialects) compared to smaller counterparts \cite{hofmann_ai_2024}. \citep{pfohl2024toolbox} perform the largest-scale health equity evaluation of large language models to date, highlighting the complexity of equity challenges and the necessity of careful, multi-dimensional evaluations to identify and mitigate them.

A first challenge is mitigating stereotypes and bias in generated text. Like human clinicians \cite{hoffman_racial_2016}, LLM-generated text has been shown to display medical stereotypes \cite{zack2024assessing,omiye2023beyond}.
For example, \cite{zack2024assessing} finds that when GPT-4 is asked to provide clinical vignettes about sarcoidosis, it generates vignettes about Black patients 97\% of the time, exaggerating the true population skew. 
Due to these embedded biases, if patients specify their demographics when asking questions to LLM chatbots, there is a risk that the LLM will overestimate the impact of race, gender, etc.~in its response. 
Similarly, when generating new clinical vignettes (e.g., for use in medical education), LLMs may over-index on demographic correlations \cite{zack2024assessing}, which would skew the knowledge of medical trainees if generated vignettes are widely used \cite{ripp_raceethnicity_2017}. 
It is likely that these issues can be mitigated through improved prompting and careful auditing of generated text. 
However, LLM stereotypes remain a key risk, both because they can be hard to detect, and because even small effect sizes can cause significant harm if the models are used at scale.
Further work is necessary to better document such patterns, to properly inform users about them, and to develop mitigating strategies. 

A second equity-related challenge is disparities in patient awareness of, and willingness to use, generative interfaces. Recent surveys show that awareness of LLMs positively correlates with formal education level and household income \cite{vogels_majority_2023}. Moreover, LLMs are more accessible to ``tech-savvy'' users \cite{weidinger2022taxonomy}, since using them requires a fast internet connection, intuition around how to best phrase prompts, etc. These factors raise the possibility that generative interfaces may create larger benefits for already-privileged patients who have fewer barriers to health access. \cite{veinot_good_2018, smith_new_2019}. To mitigate this risk, we need to study the factors underlying generative interface literacy \cite{long_what_2020} and ensure they become broadly accessible.

However, generative models also open important new health-equity-related \emph{opportunities}~\cite{pierson2023use,rodriguez2024leveraging}.~\cite{pierson2023use} describes several such use cases for generative models: detecting human biases (e.g., from clinical notes); creating structured equity-relevant databases from unstructured text; and improving equity of access to health information. To identify such equity-related opportunities, it is imperative to focus on equity from the very beginning of a project, at the \emph{problem selection} stage~\cite{chen2021ethical,kim2024development}.

\subsection{Performing real-world evaluation}\label{sec:evaluation}
In order to reason about the real-world efficacy of generative models, we need fine-grained, real-world evaluations.  
Recent work has highlighted how evaluating clinical generative models on the basis of diagnosis alone overestimates their efficacy \cite{hager_evaluation_2024}. Specifically, \cite{hager_evaluation_2024} highlight the importance of quantifying the extent to which generative models fail to adhere to treatment guidelines, or are sensitive to the order in which information is presented to them; measures of these types of behavior are important towards reasoning about the impact of generative models in medicine.

The principles of fine-grained real-world evaluation apply to all use cases we highlight. For example, evaluations of generative models for medical question-answering must specify both (1) a set of prompts to test and (2) a set of criteria for a response to be deemed high-quality.
For example, on (1), common medical LLM benchmarks use questions from exams \cite{jin_what_2020, pal_medmcqa_2022}, clinical guidebooks \cite{chen_use_2023, pan_assessment_2023}, or research papers \cite{jin_pubmedqa_2019}. These evaluation sets may be systematically different from the real distribution of patient questions, both in medical content and linguistic characteristics (e.g., language, grammar, or dialect), which could affect performance \cite{lai_chatgpt_2023, deas_evaluation_2023, ghosh2023chatgpt}. More recently, evaluations have included real patient questions from forums like MedlinePlus \cite{abacha_bridging_2019, singhal_towards_2023}, Reddit (\url{/r/AskDocs}; \cite{ayers2023comparing}), or cancer support groups on Facebook \cite{yeo2023assessing}.
These results may better reflect real-world patient questions, but future work should explicitly recruit patients from underserved groups, whose needs may differ from those represented by the dominant voices on online medical forums.
On point (2), response quality is usually judged by physicians \cite{singhal_towards_2023, ayers2023comparing, ayers2023evaluating, yeo2023assessing}, not patients. 
While physicians can best evaluate correctness, patients themselves may be better judges of perceived qualities like empathy \cite{ayers2023comparing} or understandability \cite{pan_assessment_2023}.
Future work should also prioritize closing the gap between real-world usage and assumptions made during evaluation. For example, the standard approach to evaluation assumes that generative models have no capacity to collect additional information; a more realistic set-up would allow a generative model to pose follow-up questions \citep{li_mediq_2024}. More broadly, a deeper understanding of the ways in which humans interact with generative interfaces will lead to a deeper understanding of generative model failures in the real world.

\subsection{Clarifying accountability}
The introduction of generative AI in healthcare raises important questions about who should be responsible for potential harms and regulation. 
Errors from generative models are inevitable, as they are with humans, but it remains unclear whether responsibility lies with the healthcare provider, the AI system developer, or the institution implementing the technology. Uncertainty surrounding liability is a key concern for healthcare providers who interact with generative interfaces~\cite{antoniak2024nlp}, and regulation should be designed to reduce this uncertainty. 
Possible paths forward range from holding the ``manufacturer" (i.e. the model developer) completely accountable for model errors to distributing  responsibility across providers, hospitals, and model developers \citep{cestonaro2023defining}.

The question of accountability is further complicated by concerns about over-reliance. If healthcare professionals rely too heavily on generative AI tools to make accurate decisions, while model developers simultaneously rely on healthcare professionals to carefully vet those decisions, accountability may be lost. Healthcare professionals who rely too heavily on generative AI tools may not only find it difficult to make accurate decisions without them, but also be unable to detect errors when the generative AI tools are incorrect.
Behavioral changes in response to the integration of AI tools are well-established \cite{bastani2024generative,vicente2023humans} and relate to the \emph{autonomy} of decision-making, a key factor in determining liability \cite{sung2020artificial}.  
In the education space, researchers have found that students who have access to generative AI tools outperform a control group, but once the generative AI tools are removed, they perform worse~\cite{bastani2024generative}. 
Similar studies have found that humans can also inherit biases from AI even when access to tools has been removed~\cite{vicente2023humans}. 
These findings have important implications for the deployment of generative models in medicine, and suggest the importance of research and regulation that clarifies accountability in human-AI collaboration.

\section{Conclusion}

Those witnessing the explosion of interest in generative models in healthcare might justly feel both excitement and trepidation. On the one hand, increased model capabilities enable many use cases benefiting clinicians, patients, trial organizers, researchers, and trainees, with the potential to transform healthcare. But realizing this potential in high-stakes healthcare settings will require addressing numerous challenges --- from centering equity, to protecting consent, to rigorously evaluating models --- to bridge the gap between medical generative models in \emph{theory} and in \emph{practice}. 
The history of technical advances in medicine suggests that we will not be able to anticipate all the impacts of generative models --- hospitals today are still adapting to the transition to EHRs, 15 years after their widespread introduction --- and that humility is warranted. But, in the face of this uncertainty, the research directions we outline offer a roadmap for addressing generative AI's shortcomings, and realizing its potential, in order to provide better healthcare for all.

\section*{DISCLOSURE STATEMENT}
Monica Agrawal is a co-founder of Layer Health and holds equity.
Other authors are not aware of any affiliations, memberships, funding, or financial holdings that
might be perceived as affecting the objectivity of this review.


\section*{ACKNOWLEDGMENTS}
This work was supported by a Google Research Scholar award, Apple Machine Learning Faculty Research Award,  NSF CAREER \#2142419 and \#2339381, NSF DGE \#2139899, a CIFAR Azrieli Global scholarship, a Gordon \& Betty Moore Foundation award, Optum, a gift to the LinkedIn-Cornell Bowers CIS Strategic Partnership, the Abby Joseph Cohen Faculty Fund, the Center for Advancing Safety of Machine Intelligence, and a Whitehead Scholar award. Any opinions, findings, and conclusions or recommendations expressed in this material are those of the authors and do not necessarily reflect the views of the funders.

%
\bibliography{bibliography,bibliography2}

\begin{thebibliography}{271}
\expandafter\ifx\csname natexlab\endcsname\relax\def\natexlab#1{#1}\fi

\bibitem{tai2024ai}
Tai-Seale M, Baxter SL, Vaida F, Walker A, Sitapati AM, et~al. 2024.
{AI-Generated Draft Replies Integrated Into Health Records and Physicians’ Electronic Communication}.
\textit{JAMA Network Open} 7(4):e246565--e246565

\bibitem{zakka2024almanac}
Zakka C, Cho J, Fahed G, Shad R, Moor M, et~al. 2024.
Almanac copilot: Towards autonomous electronic health record navigation.
\textit{arXiv preprint arXiv:2405.07896}

\bibitem{garcia2024artificial}
Garcia P, Ma SP, Shah S, Smith M, Jeong Y, et~al. 2024.
Artificial intelligence--generated draft replies to patient inbox messages.
\textit{JAMA Network Open} 7(3):e243201--e243201

\bibitem{kamb2024}
Kambhamettu H, Metaxa D, Johnson K, Head A. 2024{\natexlab{a}}.
\textit{Explainable Notes: Examining How to Unlock Meaning in Medical Notes with Interactivity and Artificial Intelligence}.
In \textit{Proceedings of the CHI Conference on Human Factors in Computing Systems}, CHI '24. New York, NY, USA: Association for Computing Machinery

\bibitem{mannhardt2024impact}
Mannhardt N, Bondi-Kelly E, Lam B, O'Connell C, Asiedu M, et~al. 2024.
Impact of large language model assistance on patients reading clinical notes: A mixed-methods study.
\textit{arXiv preprint arXiv:2401.09637}

\bibitem{tierney2024ambient}
Tierney AA, Gayre G, Hoberman B, Mattern B, Ballesca M, et~al. 2024.
Ambient artificial intelligence scribes to alleviate the burden of clinical documentation.
\textit{NEJM Catalyst Innovations in Care Delivery} 5(3):CAT--23

\bibitem{marshall2020trialstreamer}
Marshall IJ, Nye B, Kuiper J, Noel-Storr A, Marshall R, et~al. 2020.
Trialstreamer: A living, automatically updated database of clinical trial reports.
\textit{Journal of the American Medical Informatics Association} 27(12):1903--1912

\bibitem{feldman_scaling_2024}
Feldman J, Hochman KA, Guzman BV, Goodman A, Weisstuch J, Testa P. 2024.
Scaling {Note} {Quality} {Assessment} {Across} an {Academic} {Medical} {Center} with {AI} and {GPT}-4.
\textit{NEJM Catalyst} 5(5):CAT.23.0283

\bibitem{liu2024using}
Liu S, Wright AP, Mccoy AB, Huang SS, Genkins JZ, et~al. 2024.
Using large language model to guide patients to create efficient and comprehensive clinical care message.
\textit{Journal of the American Medical Informatics Association} :ocae142

\bibitem{wang_autotrial_2023}
Wang Z, Xiao C, Sun J. 2023.
\textit{{AutoTrial}: {Prompting} {Language} {Models} for {Clinical} {Trial} {Design}}.
In \textit{Proceedings of the 2023 {Conference} on {Empirical} {Methods} in {Natural} {Language} {Processing}}, ed. H~Bouamor, J~Pino, K~Bali, pp.  12461--12472, pp.  12461--12472. Singapore: Association for Computational Linguistics

\bibitem{white_clinidigest_2023}
White RD, Peng T, Sripitak P, Johansen AR, Snyder M. 2023.
\textit{{CliniDigest}: {A} {Case} {Study} in {Large} {Language} {Model} {Based} {Large}-{Scale} {Summarization} of {Clinical} {Trial} {Descriptions}}.
In \textit{Proceedings of the 2023 {ACM} {Conference} on {Information} {Technology} for {Social} {Good}}, pp.  396--402

\bibitem{guo_automated_2024}
Guo E, Gupta M, Deng J, Park YJ, Paget M, Naugler C. 2024.
Automated {Paper} {Screening} for {Clinical} {Reviews} {Using} {Large} {Language} {Models}: {Data} {Analysis} {Study}.
\textit{Journal of Medical Internet Research} 26:e48996

\bibitem{ktena_generative_2024}
Ktena I, Wiles O, Albuquerque I, Rebuffi SA, Tanno R, et~al. 2024.
Generative models improve fairness of medical classifiers under distribution shifts.
\textit{Nature Medicine} 30(4):1166--1173

\bibitem{bakkum_using_2024}
Bakkum MJ, Hartjes MG, Piët JD, Donker EM, Likic R, et~al. 2024.
Using artificial intelligence to create diverse and inclusive medical case vignettes for education.
\textit{British Journal of Clinical Pharmacology} 90(3):640--648

\bibitem{dupont2021decade}
Dupont PE, Nelson BJ, Goldfarb M, Hannaford B, Menciassi A, et~al. 2021.
A decade retrospective of medical robotics research from 2010 to 2020.
\textit{Science robotics} 6(60):eabi8017

\bibitem{kononenko2001machine}
Kononenko I. 2001.
Machine learning for medical diagnosis: history, state of the art and perspective.
\textit{Artificial Intelligence in medicine} 23(1):89--109

\bibitem{razzak2020big}
Razzak MI, Imran M, Xu G. 2020.
Big data analytics for preventive medicine.
\textit{Neural Computing and Applications} 32(9):4417--4451

\bibitem{kruse2017security}
Kruse CS, Smith B, Vanderlinden H, Nealand A. 2017.
Security techniques for the electronic health records.
\textit{Journal of medical systems} 41:1--9

\bibitem{fernandez2013security}
Fern{\'a}ndez-Alem{\'a}n JL, Se{\~n}or IC, Lozoya P{\'A}O, Toval A. 2013.
Security and privacy in electronic health records: A systematic literature review.
\textit{Journal of biomedical informatics} 46(3):541--562

\bibitem{campanella2016impact}
Campanella P, Lovato E, Marone C, Fallacara L, Mancuso A, et~al. 2016.
The impact of electronic health records on healthcare quality: a systematic review and meta-analysis.
\textit{The European Journal of Public Health} 26(1):60--64

\bibitem{graber2019electronic}
Graber ML, Siegal D, Riah H, Johnston D, Kenyon K. 2019.
Electronic health record--related events in medical malpractice claims.
\textit{Journal of patient safety} 15(2):77--85

\bibitem{ancker2017effects}
Ancker JS, Edwards A, Nosal S, Hauser D, Mauer E, et~al. 2017.
Effects of workload, work complexity, and repeated alerts on alert fatigue in a clinical decision support system.
\textit{BMC medical informatics and decision making} 17:1--9

\bibitem{rosenberg1982learning}
Rosenberg N. 1982.
Learning by using.
\textit{Inside the black box: Technology and economics} :120--140

\bibitem{bishop2006pattern}
Bishop CM, Nasrabadi NM. 2006.
\textit{Pattern recognition and machine learning}, vol.~4.
Springer

\bibitem{kaplan_scaling_2020a}
Kaplan J, McCandlish S, Henighan T, Brown TB, Chess B, et~al. 2020.
Scaling {Laws} for {Neural} {Language} {Models}.
\textit{arXiv:2001.08361 [cs, stat]} ArXiv: 2001.08361

\bibitem{movva2024topics}
Movva R, Balachandar S, Peng K, Agostini G, Garg N, Pierson E. 2024.
\textit{Topics, Authors, and Institutions in Large Language Model Research: Trends from 17K arXiv Papers}.
In \textit{Proceedings of the 2024 Conference of the North American Chapter of the Association for Computational Linguistics: Human Language Technologies (Volume 1: Long Papers)}, pp.  1223--1243

\bibitem{abbaspourazad_large-scale_2023}
Abbaspourazad S, Elachqar O, Miller AC, Emrani S, Nallasamy U, Shapiro I. 2023.
Large-scale {Training} of {Foundation} {Models} for {Wearable} {Biosignals}

\bibitem{beaini_towards_2023}
Beaini D, Huang S, Cunha JA, Li Z, Moisescu-Pareja G, et~al. 2023.
Towards {Foundational} {Models} for {Molecular} {Learning} on {Large}-{Scale} {Multi}-{Task} {Datasets}

\bibitem{mckeen_ecg-fm_2024}
McKeen K, Oliva L, Masood S, Toma A, Rubin B, Wang B. 2024.
{ECG}-{FM}: {An} {Open} {Electrocardiogram} {Foundation} {Model}

\bibitem{bond-taylor_deep_2021}
Bond-Taylor S, Leach A, Long Y, Willcocks CG. 2021.
Deep {Generative} {Modelling}: {A} {Comparative} {Review} of {VAEs}, {GANs}, {Normalizing} {Flows}, {Energy}-{Based} and {Autoregressive} {Models}

\bibitem{vaswani2017attention}
Vaswani A, Shazeer N, Parmar N, Uszkoreit J, Jones L, et~al. 2017.
Attention is all you need.
\textit{Advances in neural information processing systems} 30

\bibitem{jurafsky2000speech}
Jurafsky D. 2000.
Speech and language processing

\bibitem{ouyang_Training_2022}
Ouyang L, Wu J, Jiang X, Almeida D, Wainwright CL, et~al. 2022.
Training language models to follow instructions with human feedback.
ArXiv:2203.02155 [cs]

\bibitem{bolton_biomedlm_2024}
Bolton E, Venigalla A, Yasunaga M, Hall D, Xiong B, et~al. 2024.
{BioMedLM}: {A} 2.{7B} {Parameter} {Language} {Model} {Trained} {On} {Biomedical} {Text}.
ArXiv:2403.18421 [cs]

\bibitem{chen_meditron70b_2023}
Chen Z, Cano AH, Romanou A, Bonnet A, Matoba K, et~al. 2023{\natexlab{a}}.
{MEDITRON}-{70B}: {Scaling} {Medical} {Pretraining} for {Large} {Language} {Models}.
ArXiv:2311.16079 [cs]

\bibitem{singhal_expertlevel_2023}
Singhal K, Tu T, Gottweis J, Sayres R, Wulczyn E, et~al. 2023{\natexlab{a}}.
Towards {Expert}-{Level} {Medical} {Question} {Answering} with {Large} {Language} {Models}.
ArXiv:2305.09617 [cs]

\bibitem{jin_what_2021}
Jin D, Pan E, Oufattole N, Weng WH, Fang H, Szolovits P. 2021.
What {Disease} {Does} {This} {Patient} {Have}? {A} {Large}-{Scale} {Open} {Domain} {Question} {Answering} {Dataset} from {Medical} {Exams}.
\textit{Applied Sciences} 11(14):6421Number: 14 Publisher: Multidisciplinary Digital Publishing Institute

\bibitem{fleming_medalign_2023}
Fleming SL, Lozano A, Haberkorn WJ, Jindal JA, Reis EP, et~al. 2023.
{MedAlign}: {A} {Clinician}-{Generated} {Dataset} for {Instruction} {Following} with {Electronic} {Medical} {Records}.
ArXiv:2308.14089 [cs]

\bibitem{rasmy_medbert_2021}
Rasmy L, Xiang Y, Xie Z, Tao C, Zhi D. 2021.
Med-{BERT}: pretrained contextualized embeddings on large-scale structured electronic health records for disease prediction.
\textit{npj Digital Medicine} 4(1):1--13Publisher: Nature Publishing Group

\bibitem{hill_chiron_2023}
Hill BL, Emami M, Nori VS, Cordova-Palomera A, Tillman RE, Halperin E. 2023.
\textit{{CHIRon}: {A} {Generative} {Foundation} {Model} for {Structured} {Sequential} {Medical} {Data}}

\bibitem{ferruz2022protgpt2}
Ferruz N, Schmidt S, H{\"o}cker B. 2022.
Protgpt2 is a deep unsupervised language model for protein design.
\textit{Nature communications} 13(1):4348

\bibitem{rives_Biological_2021}
Rives A, Meier J, Sercu T, Goyal S, Lin Z, et~al. 2021.
Biological structure and function emerge from scaling unsupervised learning to 250 million protein sequences.
\textit{Proceedings of the National Academy of Sciences} 118(15):e2016239118Publisher: Proceedings of the National Academy of Sciences

\bibitem{nguyen_sequence_2024}
Nguyen E, Poli M, Durrant MG, Thomas AW, Kang B, et~al. 2024.
Sequence modeling and design from molecular to genome scale with {Evo}.
Pages: 2024.02.27.582234 Section: New Results

\bibitem{yang_diffusion_2023}
Yang L, Zhang Z, Song Y, Hong S, Xu R, et~al. 2023.
Diffusion {Models}: {A} {Comprehensive} {Survey} of {Methods} and {Applications}.
\textit{ACM Comput. Surv.} 56(4):105:1--105:39

\bibitem{dhariwal_diffusion_2021}
Dhariwal P, Nichol A. 2021.
Diffusion {Models} {Beat} {GANs} on {Image} {Synthesis}.
ArXiv:2105.05233 [cs, stat]

\bibitem{kazerouni_diffusion_2023}
Kazerouni A, Aghdam EK, Heidari M, Azad R, Fayyaz M, et~al. 2023.
Diffusion models in medical imaging: {A} comprehensive survey.
\textit{Medical Image Analysis} 88:102846

\bibitem{sun_aligning_2023a}
Sun S, Goldgof G, Butte A, Alaa AM. 2023.
Aligning {Synthetic} {Medical} {Images} with {Clinical} {Knowledge} using {Human} {Feedback}.
\textit{Advances in Neural Information Processing Systems} 36:13408--13428

\bibitem{radford_learning_2021}
Radford A, Kim JW, Hallacy C, Ramesh A, Goh G, et~al. 2021.
\textit{Learning {Transferable} {Visual} {Models} {From} {Natural} {Language} {Supervision}}.
In \textit{Proceedings of the 38th {International} {Conference} on {Machine} {Learning}}, pp.  8748--8763. PMLR.
ISSN: 2640-3498

\bibitem{chambon_adapting_2022}
Chambon PJM, Bluethgen C, Langlotz C, Chaudhari A. 2022.
\textit{Adapting {Pretrained} {Vision}-{Language} {Foundational} {Models} to {Medical} {Imaging} {Domains}}

\bibitem{li_BLIP2_2023}
Li J, Li D, Savarese S, Hoi S. 2023{\natexlab{a}}.
\textit{{BLIP}-2: {Bootstrapping} {Language}-{Image} {Pre}-training with {Frozen} {Image} {Encoders} and {Large} {Language} {Models}}.
In \textit{Proceedings of the 40th {International} {Conference} on {Machine} {Learning}}, pp.  19730--19742. PMLR.
ISSN: 2640-3498

\bibitem{thawakar_xraygpt_2024}
Thawakar OC, Shaker AM, Mullappilly SS, Cholakkal H, Anwer RM, et~al. 2024.
\textit{{XrayGPT}: {Chest} {Radiographs} {Summarization} using {Large} {Medical} {Vision}-{Language} {Models}}.
In \textit{Proceedings of the 23rd {Workshop} on {Biomedical} {Natural} {Language} {Processing}}, ed. D~Demner-Fushman, S~Ananiadou, M~Miwa, K~Roberts, J~Tsujii, pp.  440--448, pp.  440--448. Bangkok, Thailand: Association for Computational Linguistics

\bibitem{bazi_vision_2023}
Bazi Y, Rahhal MMA, Bashmal L, Zuair M. 2023.
Vision–{Language} {Model} for {Visual} {Question} {Answering} in {Medical} {Imagery}.
\textit{Bioengineering} 10(3):380Number: 3 Publisher: Multidisciplinary Digital Publishing Institute

\bibitem{li_llavamed_2023}
Li C, Wong C, Zhang S, Usuyama N, Liu H, et~al. 2023{\natexlab{b}}.
{LLaVA}-{Med}: {Training} a {Large} {Language}-and-{Vision} {Assistant} for {Biomedicine} in {One} {Day}.
\textit{Advances in Neural Information Processing Systems} 36:28541--28564

\bibitem{aiken_physician_2023}
Aiken LH, Lasater KB, Sloane DM, Pogue CA, Fitzpatrick~Rosenbaum KE, et~al. 2023.
Physician and {Nurse} {Well}-{Being} and {Preferred} {Interventions} to {Address} {Burnout} in {Hospital} {Practice}: {Factors} {Associated} {With} {Turnover}, {Outcomes}, and {Patient} {Safety}.
\textit{JAMA Health Forum} 4(7):e231809

\bibitem{saag2019pajama}
Saag HS, Shah K, Jones SA, Testa PA, Horwitz LI. 2019.
Pajama time: working after work in the electronic health record.
\textit{Journal of general internal medicine} 34:1695--1696

\bibitem{heer2019agency}
Heer J. 2019.
Agency plus automation: Designing artificial intelligence into interactive systems.
\textit{Proceedings of the National Academy of Sciences} 116(6):1844--1850

\bibitem{mamykina2012clinical}
Mamykina L, Vawdrey DK, Stetson PD, Zheng K, Hripcsak G. 2012.
Clinical documentation: composition or synthesis?
\textit{Journal of the American Medical Informatics Association} 19(6):1025--1031

\bibitem{jiang2023conceptualizing}
Jiang S, Shen S, Agrawal M, Lam B, Kurtzman N, et~al. 2023.
\textit{Conceptualizing machine learning for dynamic information retrieval of electronic health record notes}.
In \textit{Machine Learning for Healthcare Conference}, pp.  343--359. PMLR

\bibitem{young_diagnostic_nodate}
Young CC, Enichen E, Rivera C, Auger CA, Grant N, et~al. 2024.
Diagnostic {Accuracy} of a {Custom} {Large} {Language} {Model} on {Rare} {Pediatric} {Disease} {Case} {Reports}.
\textit{American Journal of Medical Genetics Part A} n/a(n/a):e63878

\bibitem{kanjee2023accuracy}
Kanjee Z, Crowe B, Rodman A. 2023{\natexlab{a}}.
Accuracy of a generative artificial intelligence model in a complex diagnostic challenge.
\textit{JAMA}

\bibitem{levine_diagnostic_2023}
Levine DM, Tuwani R, Kompa B, Varma A, Finlayson SG, et~al. 2023.
The {Diagnostic} and {Triage} {Accuracy} of the {GPT}-3 {Artificial} {Intelligence} {Model}

\bibitem{rios-hoyo_evaluation_2024}
Ríos-Hoyo A, Shan NL, Li A, Pearson AT, Pusztai L, Howard FM. 2024.
Evaluation of large language models as a diagnostic aid for complex medical cases.
\textit{Frontiers in Medicine} 11

\bibitem{olmo_assessing_2024}
Olmo Jd, Logroño J, Mascías C, Martínez M, Isla J. 2024.
Assessing {DxGPT}: {Diagnosing} {Rare} {Diseases} with {Various} {Large} {Language} {Models}

\bibitem{kanjee_accuracy_2023}
Kanjee Z, Crowe B, Rodman A. 2023{\natexlab{b}}.
Accuracy of a {Generative} {Artificial} {Intelligence} {Model} in a {Complex} {Diagnostic} {Challenge}.
\textit{JAMA} 330(1):78--80

\bibitem{zhou_pre-trained_2024}
Zhou J, He X, Sun L, Xu J, Chen X, et~al. 2024{\natexlab{a}}.
Pre-trained multimodal large language model enhances dermatological diagnosis using {SkinGPT}-4.
\textit{Nature Communications} 15(1):5649

\bibitem{thawkar_xraygpt_2023}
Thawkar O, Shaker A, Mullappilly SS, Cholakkal H, Anwer RM, et~al. 2023.
{XrayGPT}: {Chest} {Radiographs} {Summarization} using {Medical} {Vision}-{Language} {Models}

\bibitem{moor_med-flamingo_2023}
Moor M, Huang Q, Wu S, Yasunaga M, Zakka C, et~al. 2023.
Med-{Flamingo}: a {Multimodal} {Medical} {Few}-shot {Learner}

\bibitem{lin_medical_2023}
Lin Z, Zhang D, Tao Q, Shi D, Haffari G, et~al. 2023.
Medical visual question answering: {A} survey.
\textit{Artificial Intelligence in Medicine} 143:102611

\bibitem{reese_limitations_2024}
Reese JT, Danis D, Caufield JH, Groza T, Casiraghi E, et~al. 2024.
On the limitations of large language models in clinical diagnosis.
\textit{medRxiv} :2023.07.13.23292613

\bibitem{hager_evaluation_2024}
Hager P, Jungmann F, Holland R, Bhagat K, Hubrecht I, et~al. 2024.
Evaluation and mitigation of the limitations of large language models in clinical decision-making.
\textit{Nature Medicine} 30(9):2613--2622

\bibitem{ahmed2011effect}
Ahmed A, Chandra S, Herasevich V, Gajic O, Pickering BW. 2011.
The effect of two different electronic health record user interfaces on intensive care provider task load, errors of cognition, and performance.
\textit{Critical care medicine} 39(7):1626--1634

\bibitem{murray2021medknowts}
Murray L, Gopinath D, Agrawal M, Horng S, Sontag D, Karger DR. 2021.
\textit{Medknowts: unified documentation and information retrieval for electronic health records}.
In \textit{The 34th Annual ACM Symposium on User Interface Software and Technology}, pp.  1169--1183

\bibitem{zheng2009interface}
Zheng K, Padman R, Johnson MP, Diamond HS. 2009.
An interface-driven analysis of user interactions with an electronic health records system.
\textit{Journal of the American Medical Informatics Association} 16(2):228--237

\bibitem{sackett1995need}
Sackett DL, Rosenberg WMC. 1995.
On the need for evidence-based medicine.
\textit{Journal of Public Health} 17(3):330--334

\bibitem{bastian2010seventy}
Bastian H, Glasziou P, Chalmers I. 2010.
Seventy-five trials and eleven systematic reviews a day: how will we ever keep up?
\textit{PLoS medicine} 7(9):e1000326

\bibitem{zakka_almanac_2023}
Zakka C, Chaurasia A, Shad R, Dalal AR, Kim JL, et~al. 2023.
Almanac: {Retrieval}-{Augmented} {Language} {Models} for {Clinical} {Medicine}.
ArXiv:2303.01229 [cs]

\bibitem{lee_seetrials_2024}
Lee K, Paek H, Huang LC, Hilton CB, Datta S, et~al. 2024.
{SEETrials}: {Leveraging} {Large} {Language} {Models} for {Safety} and {Efficacy} {Extraction} in {Oncology} {Clinical} {Trials}

\bibitem{chewning2012patient}
Chewning B, Bylund CL, Shah B, Arora NK, Gueguen JA, Makoul G. 2012.
Patient preferences for shared decisions: a systematic review.
\textit{Patient education and counseling} 86(1):9--18

\bibitem{adler2022developing}
Adler RF, Morales P, Sotelo J, Magasi S. 2022.
Developing an mhealth app for empowering cancer survivors with disabilities: Co-design study.
\textit{JMIR Formative Research} 6(7):e37706

\bibitem{noack2021designing}
Noack EM, Schulze J, M{\"u}ller F. 2021.
Designing an app to overcome language barriers in the delivery of emergency medical services: participatory development process.
\textit{JMIR mHealth and uHealth} 9(4):e21586

\bibitem{danieli2021conversational}
Danieli M, Ciulli T, Mousavi SM, Riccardi G. 2021.
A conversational artificial intelligence agent for a mental health care app: Evaluation study of its participatory design.
\textit{JMIR Formative Research} 5(12):e30053

\bibitem{martin2019exploring}
Martin-Hammond A, Vemireddy S, Rao K, et~al. 2019.
Exploring older adults’ beliefs about the use of intelligent assistants for consumer health information management: A participatory design study.
\textit{JMIR aging} 2(2):e15381

\bibitem{amante2015access}
Amante DJ, Hogan TP, Pagoto SL, English TM, Lapane KL. 2015.
Access to care and use of the internet to search for health information: results from the us national health interview survey.
\textit{Journal of medical Internet research} 17(4):e106

\bibitem{thapa2021influence}
Thapa DK, Visentin DC, Kornhaber R, West S, Cleary M. 2021.
The influence of online health information on health decisions: A systematic review.
\textit{Patient education and counseling} 104(4):770--784

\bibitem{vanessa_choy_can_2024}
{Vanessa Choy}, {Sara Martin}, {Ashley Lumpkin}. 2024.
Can we rely on generative {AI} for healthcare information? {\textbar} {Ipsos}

\bibitem{alex_montero_kff_nodate}
{Alex Montero}, {Grace Sparks}, {Marley Presiado}, {Liz Hamel}. 2024.
{KFF} {Health} {Misinformation} {Tracking} {Poll}: {Health} and {Election} {Issues} on {TikTok} {\textbar} {KFF}

\bibitem{hersh2024search}
Hersh W. 2024.
Search still matters: information retrieval in the era of generative ai.
\textit{Journal of the American Medical Informatics Association} :ocae014

\bibitem{grossman2018leveraging}
Grossman LV, Feiner SK, Mitchell EG, Creber RMM. 2018{\natexlab{a}}.
Leveraging patient-reported outcomes using data visualization.
\textit{Applied clinical informatics} 9(03):565--575

\bibitem{zhao2017barriers}
Zhao JY, Song B, Anand E, Schwartz D, Panesar M, et~al. 2017.
\textit{Barriers, facilitators, and solutions to optimal patient portal and personal health record use: a systematic review of the literature}.
In \textit{AMIA annual symposium proceedings}, vol. 2017, pp.  1913. American Medical Informatics Association

\bibitem{grossman2018implementation}
Grossman LV, Choi SW, Collins S, Dykes PC, O’Leary KJ, et~al. 2018{\natexlab{b}}.
Implementation of acute care patient portals: recommendations on utility and use from six early adopters.
\textit{Journal of the American Medical Informatics Association} 25(4):370--379

\bibitem{grossman2019interventions}
Grossman LV, Masterson~Creber RM, Benda NC, Wright D, Vawdrey DK, Ancker JS. 2019.
Interventions to increase patient portal use in vulnerable populations: a systematic review.
\textit{Journal of the American Medical Informatics Association} 26(8-9):855--870

\bibitem{warren2019working}
Warren LR, Harrison M, Arora S, Darzi A. 2019.
Working with patients and the public to design an electronic health record interface: a qualitative mixed-methods study.
\textit{BMC medical informatics and decision making} 19:1--8

\bibitem{kambhamettu2024explainable}
Kambhamettu H, Metaxa D, Johnson K, Head A. 2024{\natexlab{b}}.
\textit{Explainable Notes: Examining How to Unlock Meaning in Medical Notes with Interactivity and Artificial Intelligence}.
In \textit{Proceedings of the CHI Conference on Human Factors in Computing Systems}, pp.  1--19

\bibitem{luo_rexplain_2024}
Luo L, Vairavamurthy J, Zhang X, Kumar A, Ter-Oganesyan RR, et~al. 2024.
{ReXplain}: {Translating} {Radiology} into {Patient}-{Friendly} {Video} {Reports}

\bibitem{basumedeasi}
Basu C, Vasu R, Yasunaga M, Yang Q. 2023.
\textit{Med-EASi: finely annotated dataset and models for controllable simplification of medical texts}.
In \textit{Proceedings of the Thirty-Seventh AAAI Conference on Artificial Intelligence and Thirty-Fifth Conference on Innovative Applications of Artificial Intelligence and Thirteenth Symposium on Educational Advances in Artificial Intelligence}, AAAI'23/IAAI'23/EAAI'23. AAAI Press

\bibitem{mirza2024using}
Mirza FN, Tang OY, Connolly ID, Abdulrazeq HA, Lim RK, et~al. 2024.
Using chatgpt to facilitate truly informed medical consent.
\textit{NEJM AI} 1(2):AIcs2300145

\bibitem{shadbolt_analysis_2023}
Shadbolt C, Naufal E, Bunzli S, Price V, Rele S, et~al. 2023.
Analysis of {Rates} of {Completion}, {Delays}, and {Participant} {Recruitment} in {Randomized} {Clinical} {Trials} in {Surgery}.
\textit{JAMA Network Open} 6(1):e2250996

\bibitem{ross_time_2013}
Ross JS, Mocanu M, Lampropulos JF, Tse T, Krumholz HM. 2013.
Time to {Publication} {Among} {Completed} {Clinical} {Trials}.
\textit{JAMA Internal Medicine} 173(9):825--828

\bibitem{zarin_update_2017}
Zarin DA, Tse T, Williams RJ, Rajakannan T. 2017.
Update on {Trial} {Registration} 11 {Years} after the {ICMJE} {Policy} {Was} {Established}.
\textit{New England Journal of Medicine} 376(4):383--391

\bibitem{getz_impact_2016}
Getz KA, Stergiopoulos S, Short M, Surgeon L, Krauss R, et~al. 2016.
The {Impact} of {Protocol} {Amendments} on {Clinical} {Trial} {Performance} and {Cost}.
\textit{Therapeutic Innovation \& Regulatory Science} 50(4):436--441

\bibitem{fogel_factors_2018}
Fogel DB. 2018.
Factors associated with clinical trials that fail and opportunities for improving the likelihood of success: {A} review.
\textit{Contemporary Clinical Trials Communications} 11:156--164

\bibitem{ghim_transforming_2023}
Ghim JL, Ahn S. 2023.
Transforming clinical trials: the emerging roles of large language models.
\textit{Translational and Clinical Pharmacology} 31(3):131--138

\bibitem{park_criteria2query_2024}
Park J, Fang Y, Ta C, Zhang G, Idnay B, et~al. 2024.
{Criteria2Query} 3.0: {Leveraging} generative large language models for clinical trial eligibility query generation.
\textit{Journal of Biomedical Informatics} 154:104649

\bibitem{lai_assessing_2024}
Lai H, Ge L, Sun M, Pan B, Huang J, et~al. 2024.
Assessing the {Risk} of {Bias} in {Randomized} {Clinical} {Trials} {With} {Large} {Language} {Models}.
\textit{JAMA Network Open} 7(5):e2412687

\bibitem{datta_autocriteria_2024}
Datta S, Lee K, Paek H, Manion FJ, Ofoegbu N, et~al. 2024.
{AutoCriteria}: a generalizable clinical trial eligibility criteria extraction system powered by large language models.
\textit{Journal of the American Medical Informatics Association} 31(2):375--385

\bibitem{yuan_criteria2query_2019}
Yuan C, Ryan PB, Ta C, Guo Y, Li Z, et~al. 2019.
{Criteria2Query}: a natural language interface to clinical databases for cohort definition.
\textit{Journal of the American Medical Informatics Association} 26(4):294--305

\bibitem{hamer_improving_2023}
Hamer DMd, Schoor P, Polak TB, Kapitan D. 2023.
Improving {Patient} {Pre}-screening for {Clinical} {Trials}: {Assisting} {Physicians} with {Large} {Language} {Models}.
ArXiv:2304.07396 [cs]

\bibitem{wornow_zero-shot_2024}
Wornow M, Lozano A, Dash D, Jindal J, Mahaffey KW, Shah NH. 2024.
Zero-{Shot} {Clinical} {Trial} {Patient} {Matching} with {LLMs}

\bibitem{jin_matching_2024}
Jin Q, Wang Z, Floudas CS, Chen F, Gong C, et~al. 2024.
Matching {Patients} to {Clinical} {Trials} with {Large} {Language} {Models}.
\textit{ArXiv} :arXiv:2307.15051v4

\bibitem{beattie_utilizing_nodate}
Beattie J, Neufeld S, Yang D, Chukwuma C, Gul A, et~al. 2024.
Utilizing {Large} {Language} {Models} for {Enhanced} {Clinical} {Trial} {Matching}: {A} {Study} on {Automation} in {Patient} {Screening}.
\textit{Cureus} 16(5):e60044

\bibitem{mccann_recruitment_2013}
McCann S, Campbell M, Entwistle V. 2013.
Recruitment to clinical trials: a meta-ethnographic synthesis of studies of reasons for participation.
\textit{Journal of Health Services Research \& Policy} 18(4):233--241

\bibitem{skea_exploring_2019}
Skea ZC, Newlands R, Gillies K. 2019.
Exploring non-retention in clinical trials: a meta-ethnographic synthesis of studies reporting participant reasons for drop out.
\textit{BMJ Open} 9(6):e021959

\bibitem{goodson_opportunities_2022}
Goodson N, Wicks P, Morgan J, Hashem L, Callinan S, Reites J. 2022.
Opportunities and counterintuitive challenges for decentralized clinical trials to broaden participant inclusion.
\textit{NPJ Digital Medicine} 5:58

\bibitem{thomas_artificial_2022}
Thomas KA, Kidziński L. 2022.
Artificial intelligence can improve patients’ experience in decentralized clinical trials.
\textit{Nature Medicine} 28(12):2462--2463

\bibitem{zhou_cancer_2019}
Zhou Q, Ratcliffe SJ, Grady C, Wang T, Mao JJ, Ulrich CM. 2019.
Cancer {Clinical} {Trial} {Patient}-{Participants}' {Perceptions} about {Provider} {Communication} and {Dropout} {Intentions}.
\textit{AJOB empirical bioethics} 10(3):190--200

\bibitem{dennstadt_title_2024}
Dennstädt F, Zink J, Putora PM, Hastings J, Cihoric N. 2024.
Title and abstract screening for literature reviews using large language models: an exploratory study in the biomedical domain.
\textit{Systematic Reviews} 13(1):158

\bibitem{chen_synthetic_2021}
Chen RJ, Lu MY, Chen TY, Williamson DFK, Mahmood F. 2021{\natexlab{a}}.
Synthetic data in machine learning for medicine and healthcare.
\textit{Nature Biomedical Engineering} 5(6):493--497

\bibitem{khosravi_synthetically_2024}
Khosravi B, Li F, Dapamede T, Rouzrokh P, Gamble CU, et~al. 2024.
Synthetically enhanced: unveiling synthetic data's potential in medical imaging research.
\textit{eBioMedicine} 104:105174

\bibitem{das_conditional_2021}
Das HP, Tran R, Singh J, Yue X, Tison G, et~al. 2021.
Conditional {Synthetic} {Data} {Generation} for {Robust} {Machine} {Learning} {Applications} with {Limited} {Pandemic} {Data}

\bibitem{ive_generation_2020}
Ive J, Viani N, Kam J, Yin L, Verma S, et~al. 2020.
Generation and evaluation of artificial mental health records for {Natural} {Language} {Processing}.
\textit{npj Digital Medicine} 3(1):1--9

\bibitem{pierson2023use}
Pierson E, Shanmugam D, Movva R, Kleinberg J, Agrawal M, et~al. 2024.
Use large language models to promote health equity.
\textit{arXiv preprint arXiv:2312.14804}

\bibitem{sun_pathasst_2024}
Sun Y, Zhu C, Zheng S, Zhang K, Sun L, et~al. 2024.
{PathAsst}: {A} {Generative} {Foundation} {AI} {Assistant} {Towards} {Artificial} {General} {Intelligence} of {Pathology}

\bibitem{lu_multimodal_2024}
Lu MY, Chen B, Williamson DFK, Chen RJ, Zhao M, et~al. 2024.
A {Multimodal} {Generative} {AI} {Copilot} for {Human} {Pathology}.
\textit{Nature} :1--3

\bibitem{huang_visual-language_2023}
Huang Z, Bianchi F, Yuksekgonul M, Montine TJ, Zou J. 2023.
A visual-language foundation model for pathology image analysis using medical {Twitter}.
\textit{Nature Medicine} 29(9):2307--2316

\bibitem{wang_scientific_2023}
Wang H, Fu T, Du Y, Gao W, Huang K, et~al. 2023.
Scientific discovery in the age of artificial intelligence.
\textit{Nature} 620(7972):47--60

\bibitem{wong_scribbleprompt_2024}
Wong HE, Rakic M, Guttag J, Dalca AV. 2024.
{ScribblePrompt}: {Fast} and {Flexible} {Interactive} {Segmentation} for {Any} {Biomedical} {Image}

\bibitem{zhou_hypothesis_2024}
Zhou Y, Liu H, Srivastava T, Mei H, Tan C. 2024{\natexlab{b}}.
Hypothesis {Generation} with {Large} {Language} {Models}

\bibitem{zhong_goal_2023}
Zhong R, Zhang P, Li S, Ahn J, Klein D, Steinhardt J. 2023.
Goal {Driven} {Discovery} of {Distributional} {Differences} via {Language} {Descriptions}

\bibitem{pham_topicgpt_2024}
Pham CM, Hoyle A, Sun S, Resnik P, Iyyer M. 2024.
{TopicGPT}: {A} {Prompt}-based {Topic} {Modeling} {Framework}

\bibitem{kamienny2022end}
Kamienny PA, d'Ascoli S, Lample G, Charton F. 2022.
End-to-end symbolic regression with transformers.
\textit{Advances in Neural Information Processing Systems} 35:10269--10281

\bibitem{tayebi_arasteh_large_2024}
Tayebi~Arasteh S, Han T, Lotfinia M, Kuhl C, Kather JN, et~al. 2024.
Large language models streamline automated machine learning for clinical studies.
\textit{Nature Communications} 15(1):1603

\bibitem{mozannar_realhumaneval_2024}
Mozannar H, Chen V, Alsobay M, Das S, Zhao S, et~al. 2024.
The {RealHumanEval}: {Evaluating} {Large} {Language} {Models}' {Abilities} to {Support} {Programmers}

\bibitem{biri_assessing_nodate}
Biri SK, Kumar S, Panigrahi M, Mondal S, Behera JK, Mondal H. 2024.
Assessing the {Utilization} of {Large} {Language} {Models} in {Medical} {Education}: {Insights} {From} {Undergraduate} {Medical} {Students}.
\textit{Cureus} 15(10):e47468

\bibitem{grigorian_implications_2023}
Grigorian A, Shipley J, Nahmias J, Nguyen N, Schwed AC, et~al. 2023.
Implications of {Using} {Chatbots} for {Future} {Surgical} {Education}.
\textit{JAMA Surgery} 158(11):1220--1222

\bibitem{lee_race_2022}
Lee CR, Gilliland KO, Beck~Dallaghan GL, Tolleson-Rinehart S. 2022.
Race, ethnicity, and gender representation in clinical case vignettes: a 20-year comparison between two institutions.
\textit{BMC Medical Education} 22(1):585

\bibitem{benoit2023chatgpt}
Benoit JR. 2023.
Chatgpt for clinical vignette generation, revision, and evaluation.
\textit{medRxiv} :2023--02

\bibitem{tejani_artificial_2023}
Tejani AS, Elhalawani H, Moy L, Kohli M, Kahn CE. 2023.
Artificial {Intelligence} and {Radiology} {Education}.
\textit{Radiology: Artificial Intelligence} 5(1):e220084

\bibitem{holderried_generative_2024}
Holderried F, Stegemann-Philipps C, Herschbach L, Moldt JA, Nevins A, et~al. 2024.
A {Generative} {Pretrained} {Transformer} ({GPT})-{Powered} {Chatbot} as a {Simulated} {Patient} to {Practice} {History} {Taking}: {Prospective}, {Mixed} {Methods} {Study}.
\textit{JMIR medical education} 10:e53961

\bibitem{dai_can_2023}
Dai W, Lin J, Jin H, Li T, Tsai YS, et~al. 2023.
\textit{Can {Large} {Language} {Models} {Provide} {Feedback} to {Students}? {A} {Case} {Study} on {ChatGPT}}.
In \textit{2023 {IEEE} {International} {Conference} on {Advanced} {Learning} {Technologies} ({ICALT})}, pp.  323--325

\bibitem{belmar_artificial_2023}
Belmar F, Gaete MI, Escalona G, Carnier M, Durán V, et~al. 2023.
Artificial intelligence in laparoscopic simulation: a promising future for large-scale automated evaluations.
\textit{Surgical Endoscopy} 37(6):4942--4946

\bibitem{zhao_surgical_2022}
Zhao Y, Wang Y, Zhang J, Liu X, Li Y, et~al. 2022.
Surgical {GAN}: {Towards} real-time path planning for passive flexible tools in endovascular surgeries.
\textit{Neurocomputing} 500:567--580

\bibitem{kirby1983informed}
Kirby MD. 1983.
Informed consent: what does it mean?
\textit{Journal of medical ethics} 9(2):69--75

\bibitem{riddick2003code}
Riddick FA. 2003.
The code of medical ethics of the american medical association

\bibitem{del2005informed}
Del~Carmen MG, Joffe S. 2005.
Informed consent for medical treatment and research: a review.
\textit{The oncologist} 10(8):636--641

\bibitem{pierson2022patients}
Pierson L, Pierson E. 2022.
Patients cannot consent to care unless they know how much it costs

\bibitem{astromske2021ethical}
Astromsk{\.e} K, Pei{\v{c}}ius E, Astromskis P. 2021.
Ethical and legal challenges of informed consent applying artificial intelligence in medical diagnostic consultations.
\textit{AI \& SOCIETY} 36:509--520

\bibitem{wilcox2023ai}
Wilcox L, Brewer R, Diaz F. 2023.
Ai consent futures: A case study on voice data collection with clinicians.
\textit{Proceedings of the ACM on Human-Computer Interaction} 7(CSCW2):1--30

\bibitem{garcia2023ethical}
Garcia~Valencia OA, Suppadungsuk S, Thongprayoon C, Miao J, Tangpanithandee S, et~al. 2023.
Ethical implications of chatbot utilization in nephrology.
\textit{Journal of Personalized Medicine} 13(9):1363

\bibitem{decker2023large}
Decker H, Trang K, Ramirez J, Colley A, Pierce L, et~al. 2023.
Large language model- based chatbot vs surgeon-generated informed consent documentation for common procedures.
\textit{JAMA Network Open} 6(10):e2336997--e2336997

\bibitem{burks2019health}
Burks AC, Keim-Malpass J. 2019.
Health literacy and informed consent for clinical trials: a systematic review and implications for nurses.
\textit{Nursing: Research and Reviews} :31--40

\bibitem{simon2003groups}
Simon C, Zyzanski SJ, Eder M, Raiz P, Kodish ED, Siminoff LA. 2003.
Groups potentially at risk for making poorly informed decisions about entry into clinical trials for childhood cancer.
\textit{Journal of Clinical Oncology} 21(11):2173--2178

\bibitem{raimann2024evaluation}
Raimann FJ, Neef V, Hennighausen MC, Zacharowski K, Flinspach AN. 2024.
Evaluation of ai chatbots for the creation of patient-informed consent sheets.
\textit{Machine Learning and Knowledge Extraction} 6(2):1145--1153

\bibitem{chen2024generative}
Chen Y, Esmaeilzadeh P. 2024.
Generative ai in medical practice: in-depth exploration of privacy and security challenges.
\textit{Journal of Medical Internet Research} 26:e53008

\bibitem{bai2021advancing}
Bai X, Wang H, Ma L, Xu Y, Gan J, et~al. 2021.
Advancing covid-19 diagnosis with privacy-preserving collaboration in artificial intelligence.
\textit{Nature Machine Intelligence} 3(12):1081--1089

\bibitem{ali2022federated}
Ali M, Naeem F, Tariq M, Kaddoum G. 2022.
Federated learning for privacy preservation in smart healthcare systems: A comprehensive survey.
\textit{IEEE journal of biomedical and health informatics} 27(2):778--789

\bibitem{xu2021federated}
Xu J, Glicksberg BS, Su C, Walker P, Bian J, Wang F. 2021.
Federated learning for healthcare informatics.
\textit{Journal of healthcare informatics research} 5:1--19

\bibitem{geiping_inverting_2020}
Geiping J, Bauermeister H, Dröge H, Moeller M. 2020.
Inverting {Gradients} -- {How} easy is it to break privacy in federated learning?

\bibitem{so_securing_2023}
So J, Ali RE, Guler B, Jiao J, Avestimehr S. 2023.
Securing {Secure} {Aggregation}: {Mitigating} {Multi}-{Round} {Privacy} {Leakage} in {Federated} {Learning}

\bibitem{huang_evaluating_2021}
Huang Y, Gupta S, Song Z, Li K, Arora S. 2021.
Evaluating {Gradient} {Inversion} {Attacks} and {Defenses} in {Federated} {Learning}

\bibitem{mullainathan2022solving}
Mullainathan S, Obermeyer Z. 2022.
Solving medicine’s data bottleneck: Nightingale open science.
\textit{Nature Medicine} 28(5):897--899

\bibitem{johnson2023mimic}
Johnson AE, Bulgarelli L, Shen L, Gayles A, Shammout A, et~al. 2023.
Mimic-iv, a freely accessible electronic health record dataset.
\textit{Scientific data} 10(1):1

\bibitem{barrett2023identifying}
Barrett C, Boyd B, Bursztein E, Carlini N, Chen B, et~al. 2023.
Identifying and mitigating the security risks of generative ai.
\textit{Foundations and Trends{\textregistered} in Privacy and Security} 6(1):1--52

\bibitem{el2022impossible}
El-Mhamdi EM, Farhadkhani S, Guerraoui R, Gupta N, Hoang LN, et~al. 2022.
On the impossible safety of large ai models.
\textit{arXiv preprint arXiv:2209.15259}

\bibitem{carlini2022quantifying}
Carlini N, Ippolito D, Jagielski M, Lee K, Tramer F, Zhang C. 2022.
Quantifying memorization across neural language models.
\textit{arXiv preprint arXiv:2202.07646}

\bibitem{huang2022large}
Huang J, Shao H, Chang KCC. 2022.
Are large pre-trained language models leaking your personal information?
\textit{arXiv preprint arXiv:2205.12628}

\bibitem{carlini2021extracting}
Carlini N, Tramer F, Wallace E, Jagielski M, Herbert-Voss A, et~al. 2021.
\textit{Extracting training data from large language models}.
In \textit{30th USENIX Security Symposium (USENIX Security 21)}, pp.  2633--2650

\bibitem{choi_generating_2018}
Choi E, Biswal S, Malin B, Duke J, Stewart WF, Sun J. 2018.
Generating {Multi}-label {Discrete} {Patient} {Records} using {Generative} {Adversarial} {Networks}

\bibitem{ghosheh2024survey}
Ghosheh GO, Li J, Zhu T. 2024.
A survey of generative adversarial networks for synthesizing structured electronic health records.
\textit{ACM Computing Surveys} 56(6):1--34

\bibitem{loong_disclosure_2013}
Loong B, Zaslavsky AM, He Y, Harrington DP. 2013.
Disclosure {Control} using {Partially} {Synthetic} {Data} for {Large}-{Scale} {Health} {Surveys}, with {Applications} to {CanCORS}.
\textit{Statistics in medicine} 32(24):4139--4161

\bibitem{bommasani2023foundation}
Bommasani R, Klyman K, Longpre S, Kapoor S, Maslej N, et~al. 2023.
The foundation model transparency index.
\textit{arXiv preprint arXiv:2310.12941}

\bibitem{bommasani2024foundationtransparency}
Bommasani R, et~al. 2024.
The foundation model transparency index v1.1 may 2024.
\textit{Stanford CRFM}

\bibitem{winkler2019association}
Winkler JK, Fink C, Toberer F, Enk A, Deinlein T, et~al. 2019.
Association between surgical skin markings in dermoscopic images and diagnostic performance of a deep learning convolutional neural network for melanoma recognition.
\textit{JAMA dermatology} 155(10):1135--1141

\bibitem{oakden2020hidden}
Oakden-Rayner L, Dunnmon J, Carneiro G, R{\'e} C. 2020.
\textit{Hidden stratification causes clinically meaningful failures in machine learning for medical imaging}.
In \textit{Proceedings of the ACM conference on health, inference, and learning}, pp.  151--159

\bibitem{zech2018confounding}
Zech JR, Badgeley MA, Liu M, Costa AB, Titano JJ, Oermann EK. 2018.
Confounding variables can degrade generalization performance of radiological deep learning models.
\textit{arXiv preprint arXiv:1807.00431}

\bibitem{gilpin2018explaining}
Gilpin LH, Bau D, Yuan BZ, Bajwa A, Specter M, Kagal L. 2018.
\textit{Explaining explanations: An overview of interpretability of machine learning}.
In \textit{2018 IEEE 5th International Conference on data science and advanced analytics (DSAA)}, pp.  80--89. IEEE

\bibitem{stiglic2020interpretability}
Stiglic G, Kocbek P, Fijacko N, Zitnik M, Verbert K, Cilar L. 2020.
Interpretability of machine learning-based prediction models in healthcare.
\textit{Wiley Interdisciplinary Reviews: Data Mining and Knowledge Discovery} 10(5):e1379

\bibitem{ghassemi2021false}
Ghassemi M, Oakden-Rayner L, Beam AL. 2021.
The false hope of current approaches to explainable artificial intelligence in health care.
\textit{The Lancet Digital Health} 3(11):e745--e750

\bibitem{bilodeau2024impossibility}
Bilodeau B, Jaques N, Koh PW, Kim B. 2024.
Impossibility theorems for feature attribution.
\textit{Proceedings of the National Academy of Sciences} 121(2):e2304406120

\bibitem{zhao2024opening}
Zhao H, Yang F, Lakkaraju H, Du M. 2024.
Opening the black box of large language models: Two views on holistic interpretability.
\textit{arXiv preprint arXiv:2402.10688}

\bibitem{singh2024rethinking}
Singh C, Inala JP, Galley M, Caruana R, Gao J. 2024.
Rethinking interpretability in the era of large language models.
\textit{arXiv preprint arXiv:2402.01761}

\bibitem{agarwal2024faithfulness}
Agarwal C, Tanneru SH, Lakkaraju H. 2024.
Faithfulness vs. plausibility: On the (un) reliability of explanations from large language models.
\textit{arXiv preprint arXiv:2402.04614}

\bibitem{u_vision-language_2023}
U N, M K, J K. 2023.
Vision-{Language} {Transformer} for {Interpretable} {Pathology} {Visual} {Question} {Answering}.
\textit{IEEE journal of biomedical and health informatics} 27(4)

\bibitem{kim2024transparent}
Kim C, Gadgil SU, DeGrave AJ, Omiye JA, Cai ZR, et~al. 2024{\natexlab{a}}.
Transparent medical image ai via an image--text foundation model grounded in medical literature.
\textit{Nature Medicine} :1--12

\bibitem{chen_use_2023}
Chen S, Kann BH, Foote MB, Aerts HJWL, Savova GK, et~al. 2023{\natexlab{b}}.
Use of {Artificial} {Intelligence} {Chatbots} for {Cancer} {Treatment} {Information}.
\textit{JAMA Oncology}

\bibitem{ji_survey_2023}
Ji Z, Lee N, Frieske R, Yu T, Su D, et~al. 2023.
Survey of {Hallucination} in {Natural} {Language} {Generation}.
\textit{ACM Computing Surveys} 55(12):248:1--248:38

\bibitem{lee2023benefits}
Lee P, Bubeck S, Petro J. 2023.
Benefits, limits, and risks of gpt-4 as an ai chatbot for medicine.
\textit{New England Journal of Medicine} 388(13):1233--1239

\bibitem{pan_assessment_2023}
Pan A, Musheyev D, Bockelman D, Loeb S, Kabarriti AE. 2023.
Assessment of {Artificial} {Intelligence} {Chatbot} {Responses} to {Top} {Searched} {Queries} {About} {Cancer}.
\textit{JAMA Oncology}

\bibitem{shuster_retrieval_2021}
Shuster K, Poff S, Chen M, Kiela D, Weston J. 2021.
Retrieval {Augmentation} {Reduces} {Hallucination} in {Conversation}

\bibitem{agrawal2022large}
Agrawal M, Hegselmann S, Lang H, Kim Y, Sontag D. 2022.
\textit{Large language models are few-shot clinical information extractors}.
In \textit{Proceedings of the 2022 Conference on Empirical Methods in Natural Language Processing}, pp.  1998--2022

\bibitem{gero_self-verification_2023}
Gero Z, Singh C, Cheng H, Naumann T, Galley M, et~al. 2023.
Self-{Verification} {Improves} {Few}-{Shot} {Clinical} {Information} {Extraction}.
ArXiv:2306.00024 [cs]

\bibitem{singhal_towards_2023}
Singhal K, Tu T, Gottweis J, Sayres R, Wulczyn E, et~al. 2023{\natexlab{b}}.
Towards {Expert}-{Level} {Medical} {Question} {Answering} with {Large} {Language} {Models}.
ArXiv:2305.09617 [cs]

\bibitem{mesko_imperative_2023}
Meskó B, Topol EJ. 2023.
The imperative for regulatory oversight of large language models (or generative {AI}) in healthcare.
\textit{npj Digital Medicine} 6(1):1--6Number: 1 Publisher: Nature Publishing Group

\bibitem{jakob_nielsen_ai_2023}
{Jakob Nielsen}. 2023.
{AI}: {First} {New} {UI} {Paradigm} in 60 {Years}.
\textit{Nielsen Norman Group}

\bibitem{sai2024generative}
Sai S, Gaur A, Sai R, Chamola V, Guizani M, Rodrigues JJ. 2024.
{Generative AI for transformative healthcare: A comprehensive study of emerging models, applications, case studies and limitations}.
\textit{IEEE Access}

\bibitem{mulia2023usability}
Mulia AP, Piri PR, Tho C. 2023.
Usability analysis of text generation by chatgpt openai using system usability scale method.
\textit{Procedia Computer Science} 227:381--388

\bibitem{giunti2024cocreating}
Giunti G, Doherty CP. 2024.
Cocreating an automated mhealth apps systematic review process with generative ai: Design science research approach.
\textit{JMIR Medical Education} 10:e48949

\bibitem{tankelevitch2024metacognitive}
Tankelevitch L, Kewenig V, Simkute A, Scott AE, Sarkar A, et~al. 2024.
\textit{The metacognitive demands and opportunities of generative AI}.
In \textit{Proceedings of the CHI Conference on Human Factors in Computing Systems}, pp.  1--24

\bibitem{dang2022prompt}
Dang H, Mecke L, Lehmann F, Goller S, Buschek D. 2022.
{How to prompt? Opportunities and challenges of zero-and few-shot learning for human-AI interaction in creative applications of generative models}.
\textit{arXiv preprint arXiv:2209.01390}

\bibitem{sun2022investigating}
Sun J, Liao QV, Muller M, Agarwal M, Houde S, et~al. 2022.
\textit{Investigating explainability of generative AI for code through scenario-based design}.
In \textit{Proceedings of the 27th International Conference on Intelligent User Interfaces}, pp.  212--228

\bibitem{subramonyam2024bridging}
Subramonyam H, Pea R, Pondoc C, Agrawala M, Seifert C. 2024.
\textit{Bridging the Gulf of Envisioning: Cognitive Challenges in Prompt Based Interactions with LLMs}.
In \textit{Proceedings of the CHI Conference on Human Factors in Computing Systems}, pp.  1--19

\bibitem{zamfirescu2023johnny}
Zamfirescu-Pereira J, Wong RY, Hartmann B, Yang Q. 2023.
\textit{Why Johnny can’t prompt: how non-AI experts try (and fail) to design LLM prompts}.
In \textit{Proceedings of the 2023 CHI Conference on Human Factors in Computing Systems}, pp.  1--21

\bibitem{abbasian2024foundation}
Abbasian M, Khatibi E, Azimi I, Oniani D, Shakeri Hossein~Abad Z, et~al. 2024.
Foundation metrics for evaluating effectiveness of healthcare conversations powered by generative ai.
\textit{NPJ Digital Medicine} 7(1):82

\bibitem{vasconcelos2023explanations}
Vasconcelos H, J{\"o}rke M, Grunde-McLaughlin M, Gerstenberg T, Bernstein MS, Krishna R. 2023.
Explanations can reduce overreliance on ai systems during decision-making.
\textit{Proceedings of the ACM on Human-Computer Interaction} 7(CSCW1):1--38

\bibitem{kostick2022ai}
Kostick-Quenet KM, Gerke S. 2022.
{AI in the hands of imperfect users}.
\textit{npj Digital Medicine} 5(1):197

\bibitem{wysocki2023assessing}
Wysocki O, Davies JK, Vigo M, Armstrong AC, Landers D, et~al. 2023.
{Assessing the communication gap between AI models and healthcare professionals: Explainability, utility and trust in AI-driven clinical decision-making}.
\textit{Artificial Intelligence} 316:103839

\bibitem{jacobs2021machine}
Jacobs M, Pradier MF, McCoy~Jr TH, Perlis RH, Doshi-Velez F, Gajos KZ. 2021{\natexlab{a}}.
How machine-learning recommendations influence clinician treatment selections: the example of antidepressant selection.
\textit{Translational psychiatry} 11(1):108

\bibitem{weisz2024design}
Weisz JD, He J, Muller M, Hoefer G, Miles R, Geyer W. 2024.
\textit{Design Principles for Generative AI Applications}.
In \textit{Proceedings of the CHI Conference on Human Factors in Computing Systems}, pp.  1--22

\bibitem{schaefer2016meta}
Schaefer KE, Chen JY, Szalma JL, Hancock PA. 2016.
A meta-analysis of factors influencing the development of trust in automation: Implications for understanding autonomy in future systems.
\textit{Human factors} 58(3):377--400

\bibitem{mertz2015annoying}
Mertz L. 2015.
From annoying to appreciated: Turning clinical decision support systems into a medical professional's best friend.
\textit{IEEE pulse} 6(5):4--9

\bibitem{greenes2018clinical}
Greenes RA, Bates DW, Kawamoto K, Middleton B, Osheroff J, Shahar Y. 2018.
Clinical decision support models and frameworks: seeking to address research issues underlying implementation successes and failures.
\textit{Journal of biomedical informatics} 78:134--143

\bibitem{wright2016analysis}
Wright A, Hickman TTT, McEvoy D, Aaron S, Ai A, et~al. 2016.
Analysis of clinical decision support system malfunctions: a case series and survey.
\textit{Journal of the American Medical Informatics Association} 23(6):1068--1076

\bibitem{khan2019improving}
Khan S, Richardson S, Liu A, Mechery V, McCullagh L, et~al. 2019.
Improving provider adoption with adaptive clinical decision support surveillance: an observational study.
\textit{JMIR human factors} 6(1):e10245

\bibitem{mann2020impact}
Mann D, Hess R, McGinn T, Richardson S, Jones S, et~al. 2020.
Impact of clinical decision support on antibiotic prescribing for acute respiratory infections: a cluster randomized implementation trial.
\textit{Journal of General Internal Medicine} 35:788--795

\bibitem{gaube2021ai}
Gaube S, Suresh H, Raue M, Merritt A, Berkowitz SJ, et~al. 2021.
{Do as AI say: susceptibility in deployment of clinical decision-aids}.
\textit{NPJ digital medicine} 4(1):31

\bibitem{gaube2023non}
Gaube S, Suresh H, Raue M, Lermer E, Koch TK, et~al. 2023.
Non-task expert physicians benefit from correct explainable ai advice when reviewing x-rays.
\textit{Scientific reports} 13(1):1383

\bibitem{jacobs2021designing}
Jacobs M, He J, F.~Pradier M, Lam B, Ahn AC, et~al. 2021{\natexlab{b}}.
\textit{Designing AI for trust and collaboration in time-constrained medical decisions: a sociotechnical lens}.
In \textit{Proceedings of the 2021 chi conference on human factors in computing systems}, pp.  1--14

\bibitem{henry2022factors}
Henry KE, Adams R, Parent C, Soleimani H, Sridharan A, et~al. 2022.
{Factors driving provider adoption of the TREWS machine learning-based early warning system and its effects on sepsis treatment timing}.
\textit{Nature medicine} 28(7):1447--1454

\bibitem{cabrera_improving_2023}
Cabrera AA, Perer A, Hong JI. 2023.
Improving {Human}-{AI} {Collaboration} {With} {Descriptions} of {AI} {Behavior}.
\textit{Proc. ACM Hum.-Comput. Interact.} 7(CSCW1):136:1--136:21

\bibitem{bansal_beyond_2019}
Bansal G, Nushi B, Kamar E, Lasecki WS, Weld DS, Horvitz E. 2019.
Beyond {Accuracy}: {The} {Role} of {Mental} {Models} in {Human}-{AI} {Team} {Performance}.
\textit{Proceedings of the AAAI Conference on Human Computation and Crowdsourcing} 7:2--11

\bibitem{epicshareSavesNurses}
 2024.
{G}en {A}{I} {S}aves {N}urses {T}ime by {D}rafting {R}esponses to {P}atient {M}essages --- epicshare.org.
\url{https://www.epicshare.org/share-and-learn/mayo-ai-message-responses}

\bibitem{obermeyer2019dissecting}
Obermeyer Z, Powers B, Vogeli C, Mullainathan S. 2019.
Dissecting racial bias in an algorithm used to manage the health of populations.
\textit{Science} 366(6464):447--453

\bibitem{gichoya2022ai}
Gichoya JW, Banerjee I, Bhimireddy AR, Burns JL, Celi LA, et~al. 2022.
Ai recognition of patient race in medical imaging: a modelling study.
\textit{The Lancet Digital Health} 4(6):e406--e414

\bibitem{daneshjou2022disparities}
Daneshjou R, Vodrahalli K, Novoa RA, Jenkins M, Liang W, et~al. 2022.
Disparities in dermatology ai performance on a diverse, curated clinical image set.
\textit{Science advances} 8(31):eabq6147

\bibitem{gervasi2022potential}
Gervasi SS, Chen IY, Smith-McLallen A, Sontag D, Obermeyer Z, et~al. 2022.
The potential for bias in machine learning and opportunities for health insurers to address it.
\textit{Health Affairs} 41(2):212--218

\bibitem{zink2023race}
Zink A, Obermeyer Z, Pierson E. 2024.
Race adjustments in clinical algorithms can help correct for racial disparities in data quality.
\textit{PNAS}

\bibitem{chen2021ethical}
Chen IY, Pierson E, Rose S, Joshi S, Ferryman K, Ghassemi M. 2021{\natexlab{b}}.
Ethical machine learning in healthcare.
\textit{Annual review of biomedical data science} 4:123--144

\bibitem{pierson2020assessing}
Pierson E. 2020.
Assessing racial inequality in covid-19 testing with bayesian threshold tests.
\textit{Extended abstract, NeurIPS ML4H}

\bibitem{wiens2019no}
Wiens J, Saria S, Sendak M, Ghassemi M, Liu VX, et~al. 2019.
Do no harm: a roadmap for responsible machine learning for health care.
\textit{Nature medicine} 25(9):1337--1340

\bibitem{pierson2024accuracy}
Pierson E. 2024.
Accuracy and equity in clinical risk prediction.
\textit{The New England Journal of Medicine} 390(2):100--102

\bibitem{seyyed2021underdiagnosis}
Seyyed-Kalantari L, Zhang H, McDermott MB, Chen IY, Ghassemi M. 2021.
Underdiagnosis bias of artificial intelligence algorithms applied to chest radiographs in under-served patient populations.
\textit{Nature medicine} 27(12):2176--2182

\bibitem{movva2023coarse}
Movva R, Shanmugam D, Hou K, Pathak P, Guttag J, et~al. 2023.
\textit{Coarse race data conceals disparities in clinical risk score performance}.
In \textit{Machine Learning for Healthcare Conference}, pp.  443--472. PMLR

\bibitem{zou2023implications}
Zou J, Gichoya JW, Ho DE, Obermeyer Z. 2023.
Implications of predicting race variables from medical images.
\textit{Science} 381(6654):149--150

\bibitem{balachandar2023domain}
Balachandar S, Garg N, Pierson E. 2024.
Domain constraints improve risk prediction when outcome data is missing.
\textit{ICLR}

\bibitem{ferryman2023considering}
Ferryman K, Mackintosh M, Ghassemi M. 2023.
{Considering biased data as informative artifacts in AI-assisted health care}.
\textit{New England Journal of Medicine} 389(9):833--838

\bibitem{shanmugam2024quantifying}
Shanmugam D, Hou K, Pierson E. 2024.
Quantifying disparities in intimate partner violence: a machine learning method to correct for underreporting.
\textit{npj Women's Health} 2(1):15

\bibitem{zink2020fair}
Zink A, Rose S. 2020.
Fair regression for health care spending.
\textit{Biometrics} 76(3):973--982

\bibitem{vyas2020hidden}
Vyas DA, Eisenstein LG, Jones DS. 2020.
Hidden in plain sight—reconsidering the use of race correction in clinical algorithms

\bibitem{mullainathan2021inequity}
Mullainathan S, Obermeyer Z. 2021.
\textit{On the inequity of predicting A while hoping for B}.
In \textit{AEA Papers and Proceedings}, vol. 111, pp.  37--42. American Economic Association 2014 Broadway, Suite 305, Nashville, TN 37203

\bibitem{pierson2021algorithmic}
Pierson E, Cutler DM, Leskovec J, Mullainathan S, Obermeyer Z. 2021.
An algorithmic approach to reducing unexplained pain disparities in underserved populations.
\textit{Nature Medicine} 27(1):136--140

\bibitem{obermeyer2021algorithmic}
Obermeyer Z, Nissan R, Stern M, Eaneff S, Bembeneck EJ, Mullainathan S. 2021.
Algorithmic bias playbook.
\textit{Center for Applied AI at Chicago Booth}

\bibitem{diao2024implications}
Diao JA, He Y, Khazanchi R, Tiako MN, Witonsky JI, et~al. 2024{\natexlab{a}}.
Implications of race adjustment in lung-function equations.
\textit{The New England journal of medicine} 390(22):2083

\bibitem{diao2024projected}
Diao JA, Shi I, Murthy VL, Buckley TA, Patel CJ, et~al. 2024{\natexlab{b}}.
Projected changes in statin and antihypertensive therapy eligibility with the aha prevent cardiovascular risk equations.
\textit{JAMA} 332(12):989--1000

\bibitem{pfohl2024toolbox}
Pfohl SR, Cole-Lewis H, Sayres R, Neal D, Asiedu M, et~al. 2024.
A toolbox for surfacing health equity harms and biases in large language models.
\textit{Nature Medicine} :1--11

\bibitem{hofmann_ai_2024}
Hofmann V, Kalluri PR, Jurafsky D, King S. 2024.
{AI} generates covertly racist decisions about people based on their dialect.
\textit{Nature} 633(8028):147--154

\bibitem{hoffman_racial_2016}
Hoffman KM, Trawalter S, Axt JR, Oliver MN. 2016.
Racial bias in pain assessment and treatment recommendations, and false beliefs about biological differences between blacks and whites.
\textit{Proceedings of the National Academy of Sciences of the United States of America} 113(16):4296--4301

\bibitem{zack2024assessing}
Zack T, Lehman E, Suzgun M, Rodriguez JA, Celi LA, et~al. 2024.
{Assessing the potential of GPT-4 to perpetuate racial and gender biases in health care: a model evaluation study}.
\textit{The Lancet Digital Health} 6(1):e12--e22

\bibitem{omiye2023beyond}
Omiye JA, Lester J, Spichak S, Rotemberg V, Daneshjou R. 2023.
Beyond the hype: large language models propagate race-based medicine.
\textit{medRxiv} :2023--07

\bibitem{ripp_raceethnicity_2017}
Ripp K, Braun L. 2017.
Race/{Ethnicity} in {Medical} {Education}: {An} {Analysis} of a {Question} {Bank} for {Step} 1 of the {United} {States} {Medical} {Licensing} {Examination}.
\textit{Teaching and Learning in Medicine} 29(2):115--122

\bibitem{vogels_majority_2023}
Vogels EA. 2023.
A majority of {Americans} have heard of {ChatGPT}, but few have tried it themselves

\bibitem{weidinger2022taxonomy}
Weidinger L, Uesato J, Rauh M, Griffin C, Huang PS, et~al. 2022.
\textit{Taxonomy of risks posed by language models}.
In \textit{Proceedings of the 2022 ACM Conference on Fairness, Accountability, and Transparency}, pp.  214--229

\bibitem{veinot_good_2018}
Veinot TC, Mitchell H, Ancker JS. 2018.
Good intentions are not enough: how informatics interventions can worsen inequality.
\textit{Journal of the American Medical Informatics Association: JAMIA} 25(8):1080--1088

\bibitem{smith_new_2019}
Smith B, Magnani JW. 2019.
New technologies, new disparities: {The} intersection of electronic health and digital health literacy.
\textit{International Journal of Cardiology} 292:280--282Publisher: Elsevier

\bibitem{long_what_2020}
Long D, Magerko B. 2020.
\textit{What is {AI} {Literacy}? {Competencies} and {Design} {Considerations}}.
In \textit{Proceedings of the 2020 {CHI} {Conference} on {Human} {Factors} in {Computing} {Systems}}, {CHI} '20, pp.  1--16. New York, NY, USA: Association for Computing Machinery

\bibitem{rodriguez2024leveraging}
Rodriguez JA, Alsentzer E, Bates DW. 2024.
Leveraging large language models to foster equity in healthcare.
\textit{Journal of the American Medical Informatics Association} :ocae055

\bibitem{kim2024development}
Kim JY, Hasan A, Kellogg KC, Ratliff W, Murray SG, et~al. 2024{\natexlab{b}}.
{Development and preliminary testing of Health Equity Across the AI Lifecycle (HEAAL): A framework for healthcare delivery organizations to mitigate the risk of AI solutions worsening health inequities}.
\textit{PLOS Digital Health} 3(5):e0000390

\bibitem{jin_what_2020}
Jin D, Pan E, Oufattole N, Weng WH, Fang H, Szolovits P. 2020.
What {Disease} does this {Patient} {Have}? {A} {Large}-scale {Open} {Domain} {Question} {Answering} {Dataset} from {Medical} {Exams}.
ArXiv:2009.13081 [cs]

\bibitem{pal_medmcqa_2022}
Pal A, Umapathi LK, Sankarasubbu M. 2022.
{MedMCQA} : {A} {Large}-scale {Multi}-{Subject} {Multi}-{Choice} {Dataset} for {Medical} domain {Question} {Answering}.
ArXiv:2203.14371 [cs]

\bibitem{jin_pubmedqa_2019}
Jin Q, Dhingra B, Liu Z, Cohen WW, Lu X. 2019.
{PubMedQA}: {A} {Dataset} for {Biomedical} {Research} {Question} {Answering}.
ArXiv:1909.06146 [cs, q-bio]

\bibitem{lai_chatgpt_2023}
Lai VD, Ngo NT, Veyseh APB, Man H, Dernoncourt F, et~al. 2023.
{ChatGPT} {Beyond} {English}: {Towards} a {Comprehensive} {Evaluation} of {Large} {Language} {Models} in {Multilingual} {Learning}.
ArXiv:2304.05613 [cs]

\bibitem{deas_evaluation_2023}
Deas N, Grieser J, Kleiner S, Patton D, Turcan E, McKeown K. 2023.
Evaluation of {African} {American} {Language} {Bias} in {Natural} {Language} {Generation}.
ArXiv:2305.14291 [cs]

\bibitem{ghosh2023chatgpt}
Ghosh S, Caliskan A. 2023.
{ChatGPT Perpetuates Gender Bias in Machine Translation and Ignores Non-Gendered Pronouns: Findings across Bengali and Five other Low-Resource Languages}.
\textit{arXiv preprint arXiv:2305.10510}

\bibitem{abacha_bridging_2019}
Abacha AB, Mrabet Y, Sharp M, Goodwin TR, Shooshan SE, Demner-Fushman D. 2019.
Bridging the {Gap} {Between} {Consumers}' {Medication} {Questions} and {Trusted} {Answers}.
\textit{Studies in Health Technology and Informatics} 264:25--29

\bibitem{ayers2023comparing}
Ayers JW, Poliak A, Dredze M, Leas EC, Zhu Z, et~al. 2023{\natexlab{a}}.
Comparing physician and artificial intelligence chatbot responses to patient questions posted to a public social media forum.
\textit{JAMA Internal Medicine}

\bibitem{yeo2023assessing}
Yeo YH, Samaan JS, Ng WH, Ting PS, Trivedi H, et~al. 2023.
Assessing the performance of {ChatGPT} in answering questions regarding cirrhosis and hepatocellular carcinoma.
\textit{medRxiv} :2023--02

\bibitem{ayers2023evaluating}
Ayers JW, Zhu Z, Poliak A, Leas EC, Dredze M, et~al. 2023{\natexlab{b}}.
Evaluating artificial intelligence responses to public health questions.
\textit{JAMA Network Open} 6(6):e2317517--e2317517

\bibitem{li_mediq_2024}
Li SS, Balachandran V, Feng S, Ilgen J, Pierson E, et~al. 2024.
{MEDIQ}: {Question}-{Asking} {LLMs} for {Adaptive} and {Reliable} {Clinical} {Reasoning}

\bibitem{antoniak2024nlp}
Antoniak M, Naik A, Alvarado CS, Wang LL, Chen IY. 2024.
\textit{NLP for Maternal Healthcare: Perspectives and Guiding Principles in the Age of LLMs}.
In \textit{The 2024 ACM Conference on Fairness, Accountability, and Transparency}, pp.  1446--1463

\bibitem{cestonaro2023defining}
Cestonaro C, Delicati A, Marcante B, Caenazzo L, Tozzo P. 2023.
Defining medical liability when artificial intelligence is applied on diagnostic algorithms: a systematic review.
\textit{Frontiers in Medicine} 10:1305756

\bibitem{bastani2024generative}
Bastani H, Bastani O, Sungu A, Ge H, Kabakc{\i} O, Mariman R. 2024.
Generative {AI} {Can} {Harm} {Learning}.
\textit{Available at SSRN} 4895486

\bibitem{vicente2023humans}
Vicente L, Matute H. 2023.
Humans inherit artificial intelligence biases.
\textit{Scientific Reports} 13(1):15737

\bibitem{sung2020artificial}
Sung JJ, Poon NC. 2020.
Artificial intelligence in gastroenterology: where are we heading?
\textit{Frontiers of medicine} 14(4):511--517

\end{thebibliography}
\bibliographystyle{ar-style3}

\end{document}